\documentclass[journal]{IEEEtran}
\ifCLASSINFOpdf
\else
\fi
\usepackage{palatino,epsfig,latexsym}
\usepackage{booktabs} 
\usepackage{subfigure,cite}
\usepackage{enumitem}
\usepackage{graphicx}
\usepackage{amsthm,amsmath}
\usepackage{amssymb}
\usepackage{algpseudocode}
\usepackage{algorithm}
\usepackage{color}
\usepackage{url}
\usepackage{makecell,multirow}
\usepackage{xcolor}
\usepackage{soul}
\usepackage{threeparttable}
\usepackage{cancel}

\definecolor{hl}{rgb}{0.75,0.75,0.75}
\sethlcolor{hl}

\theoremstyle{definition}
\newtheorem{definition}{Definition}

\begin{document}

\title{An Efficient Evolutionary Algorithm for Few-for-Many Optimization}
\author{Ke~Shang,~\IEEEmembership{Senior Member,~IEEE},~Hisao~Ishibuchi,~\IEEEmembership{Fellow,~IEEE},~Zexuan~Zhu,~\IEEEmembership{Senior Member,~IEEE},~Qingfu~Zhang,~\IEEEmembership{Fellow,~IEEE}
\thanks{This work was supported by National Natural Science Foundation of China (Grant No. 62472292, 62471310, 62376115), Guangdong Basic and Applied Basic Research Foundation (Grant No. 2025A1515011638), the Research Grants Council of the Hong Kong Special Administrative Region, China (GRF Project No. CityU11215622). \textit{(Corresponding Author: Ke Shang.)}}
\thanks{Ke Shang and Zexuan Zhu are with School of Artificial Intelligence, Shenzhen University, Shenzhen 518060, China (e-mail: kshang@foxmail.com;  zhuzx@szu.edu.cn). Hisao Ishibuchi is with Department of Computer Science and Engineering, Southern University of Science and Technology, Shenzhen 518055, China (e-mail: hisao@sustech.edu.cn).  Qingfu Zhang is with the Department of Computer Science, City University of Hong Kong, Hong Kong SAR, China (email: qingfu.zhang@cityu.edu.hk).}}

\maketitle

\begingroup
\renewcommand\thefootnote{}
\footnotetext{%
© 2026 IEEE. Personal use of this material is permitted. 
Permission from IEEE must be obtained for all other uses, 
in any current or future media, including reprinting/republishing 
this material for advertising or promotional purposes, creating new 
collective works, for resale or redistribution to servers or lists, 
or reuse of any copyrighted component of this work in other works.
}
\endgroup

\begin{abstract}Few-for-many (F4M) optimization, recently introduced as a novel paradigm in multi-objective optimization, aims to find a small set of solutions that effectively handle a large number of conflicting objectives. Unlike traditional many-objective optimization methods, which typically attempt comprehensive coverage of the Pareto front, F4M optimization emphasizes finding a small representative solution set to efficiently address high-dimensional objective spaces. Motivated by the computational complexity and practical relevance of F4M optimization, this paper proposes a new evolutionary algorithm explicitly tailored for efficiently solving F4M optimization problems. Inspired by SMS-EMOA, our proposed approach employs a $(\mu+1)$-evolution strategy guided by the objective of F4M optimization. Furthermore, to facilitate rigorous performance assessment, we propose a novel benchmark test suite specifically designed for F4M optimization by leveraging the similarity between the R2 indicator and F4M formulations. Our test suite is highly flexible, allowing any existing multi-objective optimization problem to be transformed into a corresponding F4M instance via scalarization using the weighted Tchebycheff function. Comprehensive experimental evaluations on benchmarks demonstrate the superior performance of our algorithm compared to existing state-of-the-art algorithms, especially on instances involving a large number of objectives. The source code of the proposed algorithm will be released publicly. Source code is available at \url{https://github.com/MOL-SZU/SoM-EMOA}.

\vspace{0.5em}
\noindent\textbf{Accepted version.} This manuscript has been accepted for publication in 
\emph{IEEE/CAA Journal of Automatica Sinica}. 
This version is the authors’ accepted manuscript and has not undergone IEEE copy-editing, 
typesetting, or final formatting.
\end{abstract}


\IEEEpeerreviewmaketitle

\section{Introduction}
\IEEEPARstart{M}{ulti}-objective optimization (MOO) has received extensive attention from the evolutionary computation community over the past decades, with a multitude of Evolutionary Multi-objective Optimization (EMO) algorithms being developed~\cite{deb2002fast,zhang2007moea,deb2013evolutionary,cheng2016reference,tian2022deep,huang2024direct}. Typically, these EMO algorithms aim to identify a diverse set of non-dominated solutions that adequately represent or approximate the Pareto front (PF) of the given multi-objective optimization problem \cite{hua2021survey}.

However, as the number of objectives increases, particularly beyond three, conventional EMO algorithms encounter severe challenges due to the increase in the dimensionality of the Pareto front~\cite{yang2013grid,wang2016diversity}. Specifically, for problems with a very large number of objectives (e.g., more than 100), the dimensionality of the Pareto front can become extremely high. Consequently, an exponentially increasing number of non-dominated solutions are required to effectively cover the entire Pareto front, leading to significant computational burdens and decision-making difficulties~\cite{ishibuchi2008evolutionary,lopez2015many,sato2023evolutionary}.

To mitigate these challenges, recent studies have proposed a new paradigm known as the few-for-many (F4M) optimization, also referred to as many-objective cover problem (MaCP)~\cite{liu2024many,lin2024few}. Unlike conventional EMO approaches, F4M optimization focuses on identifying a compact set of solutions, with size considerably smaller than the number of objectives (e.g., fewer than 10 solutions for over 100 objectives), that collectively and complementarily address all objectives. Each objective is thus approximately optimized by at least one solution within the compact set.

The rationale behind this paradigm lies in the observation that high-dimensional objective spaces often exhibit \emph{structural redundancy} and \emph{correlations among objectives}~\cite{ishibuchi2014behavior,yuan2017objective}. 
In many real-world many-objective problems, objectives are not mutually independent; instead, groups of objectives share similar trade-off patterns or can be simultaneously optimized by the same design region.
Therefore, instead of maintaining an exponentially large number of solutions to approximate the entire Pareto front, it is often sufficient to select a small number of \emph{representative and complementary solutions} that together cover these correlated objective regions~\cite{liu2024many,lin2024few,li2024many}.

The F4M optimization is highly relevant to numerous practical scenarios. For instance, F4M optimization finds applications in fields such as drug design, where a small set of drug candidates must collectively meet diverse pharmacological criteria~\cite{angelo2023multi}, and personalized advertisement systems aiming to serve diverse audience segments efficiently using a minimal set of advertisement versions~\cite{matz2017psychological,eckles2016estimating}. Further, in data mining and machine learning contexts, it is often beneficial to build a minimal set of models that collectively perform well on a diverse array of tasks~\cite{standley2020tasks,wu2022motley}.

Despite its practical significance, F4M optimization remains relatively understudied. To the best of our knowledge, existing contributions to F4M optimization are limited. Liu et al.~\cite{liu2024many} formally introduced the many-objective cover problem (MaCP), proved its NP-hard nature, and proposed a clustering-based swarm optimizer (CluSO) along with a decoupling many-objective test suite (DC-MaTS) for evaluation. Concurrently, Lin et al.~\cite{lin2024few} introduced a Tchebycheff set scalarization method (TCH-Set scalarization), extending traditional scalarization techniques to handle a small set of solutions for a large number of objectives efficiently, demonstrating promising theoretical and practical advantages. {Together, these works mark important initial steps toward formalizing and addressing the F4M paradigm, but they also reveal the substantial gap that remains in developing general benchmark frameworks and effective evolutionary algorithms tailored specifically for F4M optimization.}



{This work makes the following main contributions to the field of F4M optimization:
\begin{enumerate}
    \item \textbf{A new evolutionary algorithm for F4M optimization.}
    We propose the \emph{Sum-of-Minimum driven Evolutionary Multi-objective Optimization Algorithm (SoM-EMOA)}, 
    a novel evolutionary algorithm that directly optimizes the set-level coverage objective.
    Unlike traditional EMO algorithms such as NSGA-II, MOEA/D, or NSGA-III, 
    which approximate the entire Pareto front using large populations,
    SoM-EMOA is designed to evolve a \emph{compact population} ($k \ll m$) 
    that maximizes collective objective coverage.
    Key innovations include:
    \begin{itemize}
        \item A \emph{set-level selection mechanism} that removes individuals 
        based on marginal loss in coverage rather than dominance or decomposition;
        \item An \emph{archive-based complementarity-driven mating strategy} 
        that biases offspring generation toward underrepresented objectives;
    \end{itemize}
    Together, these designs operationalize the F4M principle within 
    a population-based search framework.
    \item \textbf{A scalable benchmark suite for F4M optimization.}
    We construct a new F4M benchmark suite derived from well-established MOPs 
    such as DTLZ and WFG by introducing an R2-based transformation 
    that expands low-dimensional Pareto fronts into many correlated scalarized objectives.
    This benchmark provides controlled scalability in both the number of objectives ($m$) 
    and the desired set size ($k$), enabling systematic evaluation of F4M algorithms.
    \item \textbf{Comprehensive empirical study and insights.}
    We conduct extensive experiments on the proposed test suite 
    and compare SoM-EMOA with representative baselines including state-of-the-art EMO algorithms, CluSO~\cite{liu2024many}, and MOCOBO~\cite{maus2025cover}.
    Results demonstrate that SoM-EMOA achieves superior coverage quality and robustness 
    in high-dimensional objective spaces.
\end{enumerate}}

The remainder of this paper is organized as follows. Section II presents the formal definition of F4M optimization and reviews related work. Section III outlines the proposed evolutionary algorithm. Section IV presents the proposed test suite for F4M optimization. Section V provides detailed descriptions of the experimental setup and discusses empirical findings. Finally, Section VI concludes the paper and outlines potential directions for future research.

\section{Preliminaries}

This section introduces the few-for-many (F4M) optimization problem. We first review the basics of multi-objective optimization and then formally define the F4M problem, followed by a brief overview of related work.

\subsection{Multi-objective Optimization}
Multi-objective optimization (MOP) aims to simultaneously optimize multiple conflicting objective functions. Mathematically, an MOP with \(m\) objectives can be formulated as follows:
\begin{equation}
\label{eq:mop}
    \min_{\boldsymbol{x} \in \mathcal{X}} \boldsymbol{f}(\boldsymbol{x}) = \left(f_{1}(\boldsymbol{x}), f_{2}(\boldsymbol{x}), \ldots, f_{m}(\boldsymbol{x})\right),
\end{equation}
where \(\mathcal{X} \subseteq \mathbb{R}^d\) denotes the decision (solution) space, and \(f_i(\boldsymbol{x}): \mathbb{R}^d \rightarrow \mathbb{R}\) is the \(i\)-th objective function. Typically, due to the conflicting nature of objectives, no single solution can simultaneously optimize all objectives~\cite{miettinen1999nonlinear,deb2016multi}. Therefore, the notion of Pareto dominance is introduced to define solution optimality:

\begin{definition}[Dominance]\label{def:dominance}
A solution \(\boldsymbol{x}^a\) is said to \emph{dominate} another solution \(\boldsymbol{x}^b\), denoted as \(\boldsymbol{x}^a \prec \boldsymbol{x}^b\), if and only if
\[
\forall i \in \{1, \dots, m\}: f_i(\boldsymbol{x}^a) \le f_i(\boldsymbol{x}^b) \quad\]
and \[\quad \exists j \in \{1,\dots,m\}: f_j(\boldsymbol{x}^a) < f_j(\boldsymbol{x}^b).
\]
\end{definition}

Based on the dominance relationship, we have the following concepts~\cite{miettinen1999nonlinear}:

\begin{definition}[Pareto Optimality] A solution \(\boldsymbol{x}^*\) is Pareto optimal if no other solution \(\boldsymbol{x}\in\mathcal{X}\) exists that dominates \(\boldsymbol{x}^*\). The set of all Pareto optimal solutions is called the Pareto optimal set (PS), and the image of PS in the objective space is called the Pareto front (PF), formally defined as follows:
\[
\text{PS} = \{ \boldsymbol{x}^*\in \mathcal{X} | \nexists \boldsymbol{x} \in \mathcal{X}: \boldsymbol{x} \prec \boldsymbol{x}^* \}, \quad \text{PF} = \{\boldsymbol{f}(\boldsymbol{x}^*) | \boldsymbol{x}^*\in \text{PS}\}.
\]
\end{definition}

\begin{figure*}[htbp]
    \centering
    \subfigure[Solution 1]{ 
\includegraphics[width=0.16\linewidth]{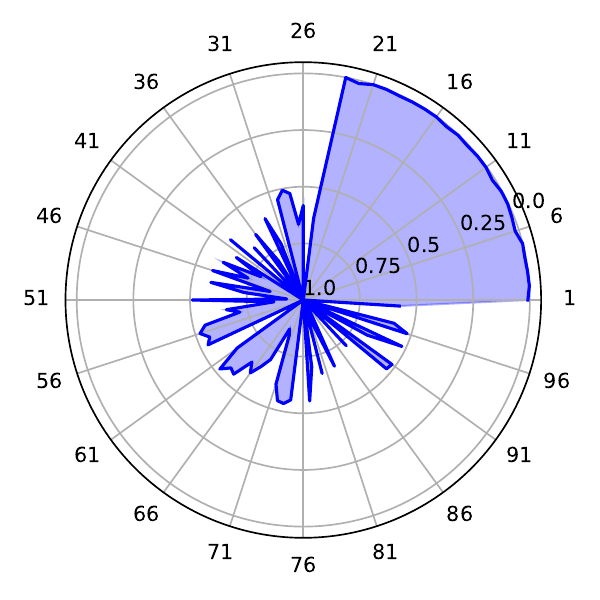}   }            
\hspace{-10pt}\subfigure[Solution 2]{ 
\includegraphics[width=0.16\linewidth]{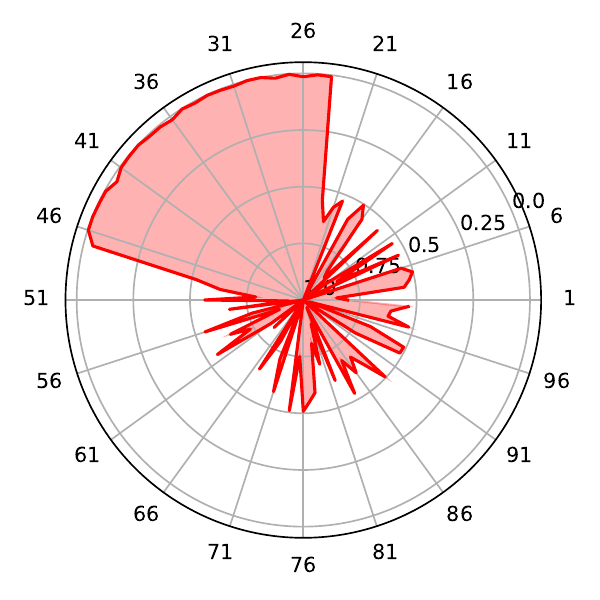} }
\hspace{-10pt}\subfigure[Solution 3]{ 
\includegraphics[width=0.16\linewidth]{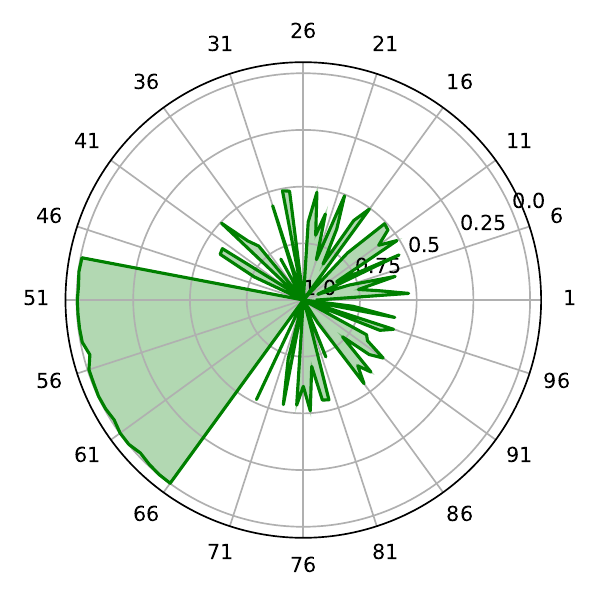} }
 \hspace{-10pt}\subfigure[Solution 4]{ 
\includegraphics[width=0.16\linewidth]{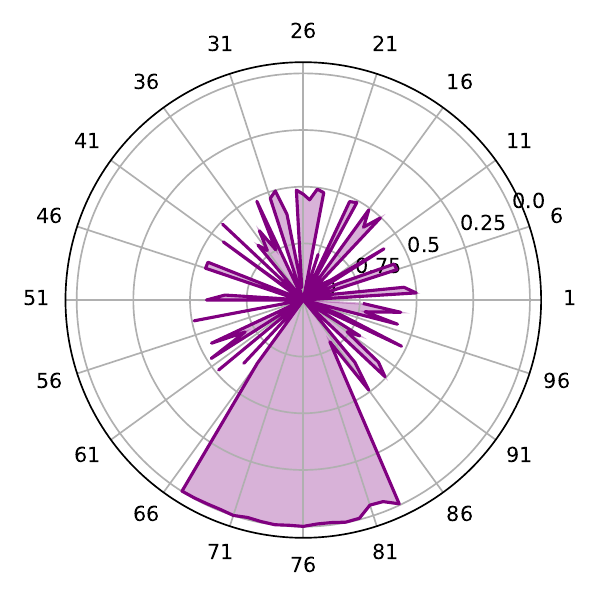} }
\hspace{-10pt}\subfigure[Solution 5]{ 
\includegraphics[width=0.16\linewidth]{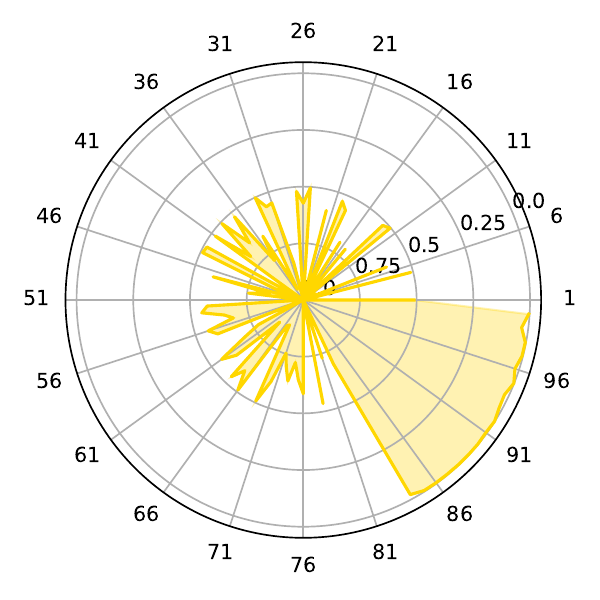} }
\hspace{-10pt}\subfigure[All solutions]{ 
\includegraphics[width=0.16\linewidth]{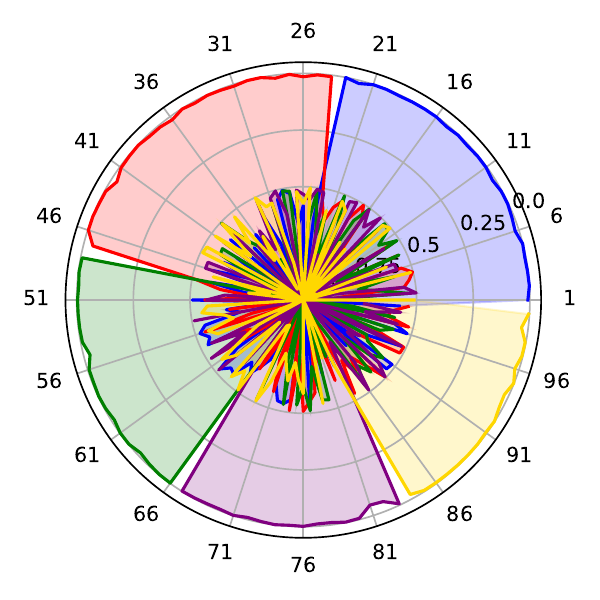} }
    \caption{An illustration of F4M optimization with \(m=100\) objectives and a cover set of size \(k=5\), {reproduced from \cite{lin2024few}}. Each polar plot represents one solution, where each angular tick corresponds to a specific objective index, and each radial tick indicates the objective value. The five solutions in (a)–(e) collectively handle different subsets of objectives in a complementary manner. Plot (f) shows the union of all five solutions, effectively handling the entire objective space.}
    \label{fig:MaCP_example}
\end{figure*}

In practice, it is usually infeasible to find the entire Pareto optimal set for real-world problems. Thus, evolutionary multi-objective optimization (EMO) algorithms such as NSGA-II~\cite{deb2002fast}, MOEA/D~\cite{zhang2007moea}, and NSGA-III~\cite{deb2013evolutionary} have been extensively studied to approximate the Pareto front by generating a finite set of diverse, representative solutions. Nevertheless, as the number of objectives increases beyond three (known as many-objective optimization, MaO), achieving good representation of the entire Pareto front becomes increasingly challenging due to computational complexity and difficulties in visualizing and interpreting the solution sets~\cite{bhattacharjee2017bridging,dy2021improving}.

\subsection{Few-for-Many Optimization}

To address the challenges of handling a large number of objectives, Liu et al.~\cite{liu2024many} recently introduced the \emph{few-for-many} (F4M) optimization paradigm, also known as the many-objective cover problem (MaCP). Instead of aiming to approximate the entire Pareto front, F4M optimization seeks a small set of solutions that \emph{collectively} perform well across a large number of objectives. 

{Formally, given $m$ objectives, the goal is to find a compact set of $k$ solutions ($k \ll m$) such that each objective is well addressed by at least one solution in the set. 
Hence, F4M optimization can be viewed as a special form of \emph{multi-objective multi-solution optimization}, where the trade-off is performed among objectives in terms of their best-attained performance across a small set of candidate solutions. Fig.~\ref{fig:MaCP_example} illustrates the basic idea of F4M optimization, where 100 objectives are handled by 5 solutions.}

\begin{definition}[Few-for-Many (F4M) Optimization]
{Given $m$ objective functions $f_1(\boldsymbol{x}), f_2(\boldsymbol{x}), \ldots, f_m(\boldsymbol{x})$ and a small integer $k \ll m$, the F4M optimization problem aims to find a solution set $X_k = \{\boldsymbol{x}^{(1)}, \boldsymbol{x}^{(2)}, \ldots, \boldsymbol{x}^{(k)}\} \subseteq \mathcal{X}$ that achieves an overall trade-off among the objectives such that each objective $f_i$ is optimized by at least one solution in $X_k$. 
Mathematically, this general problem can be formulated as:
\begin{equation}
\label{eq:f4m}
    \min_{X_k \subseteq \mathcal{X}} \; \boldsymbol{f}(X_k) = 
    \big( f_1(X_k), f_2(X_k), \ldots, f_m(X_k) \big),
\end{equation}
where $f_i(X_k)=\min_{1 \le j \le k} f_i(\boldsymbol{x}^{(j)})$ denotes the $i$th set-level objective function that reflects the best performance of $X_k$ on objective $f_i$.}
\end{definition}

{When the cardinality $|X_k|=1$, the standard F4M optimization in Eq. \eqref{eq:f4m} degenerates to the standard MOP in Eq. \eqref{eq:mop}. In other words, the classical MOP can be viewed as a special case of F4M optimization with a single representative solution.}

{Analogous to standard MOPs, the F4M problem in Eq.~\eqref{eq:f4m} can be transformed into a single-objective optimization problem through various scalarization strategies. Two representative formulations include:
\begin{itemize}
    \item \textbf{Weighted sum}~\cite{liu2024many}: 
    \begin{equation}
     \min_{X_k \subseteq \mathcal{X}} G_{\text{ws}}(X_k) =  \sum_{i=1}^{m} f_i(X_k) =
        \sum_{i=1}^{m} \min_{1 \le j \le k} f_i(\boldsymbol{x}^{(j)}).
        \label{eq:sumofmin}
    \end{equation}
        This formulation, also known as the \emph{sum-of-minimum (SoM)} approach, has been widely adopted due to its simplicity and interpretability.
    \item \textbf{Tchebycheff  scalarization}~\cite{lin2024few}: 
    \begin{equation}
    \begin{aligned}
         \min_{X_k \subseteq \mathcal{X}} G_{\text{tch}}(X_k) &= 
        \max_{1\leq i\leq m}\left\{ \lambda_i\left(f_i(X_k)-z_i^*\right)\right\}\\
        &=\max_{1\leq i\leq m}\left\{ \lambda_i\left(\min_{1 \le j \le k} f_i(\boldsymbol{x}^{(j)})-z_i^*\right)\right\},
    \end{aligned}
    \end{equation}
        where $\lambda_i > 0$ denotes a weight coefficient and $z_i^*$ is the reference (ideal) value of the $i$th objective. 
\end{itemize}}

{From this perspective, the SoM formulation (Eq.~\eqref{eq:sumofmin}) can be regarded as a specific instance of F4M optimization.
In this work, we adopt this form due to its simplicity, interpretability, and strong connection with the classical covering problem.
Nevertheless, other formulations such as  Tchebycheff scalarization represent promising alternatives for future research.}

F4M optimization generalizes and differs from classical covering problems (e.g., set cover \cite{contardo2021exact} and vertex cover \cite{zhang2022applying}) by extending the notion of covering from discrete structures to continuous optimization problems with multiple conflicting objectives. Liu et al.~\cite{liu2024many} proved that F4M optimization is an NP-hard problem by reducing it to the discrete clustering problem, highlighting the computational challenge involved.

F4M optimization provides a novel perspective on MaO by directly optimizing a small number of solutions collaboratively, thus offering a practical and computationally efficient alternative to the conventional many-objective optimization strategies that rely on dense coverage of the Pareto front.

\subsection{Related Work on F4M Optimization}
Several recent studies have addressed the problem of finding a small set of solutions to effectively handle a large number of objectives, a setting where the number of objectives significantly exceeds the number of desired solutions. These studies can be broadly grouped into three categories: evolutionary algorithms, surrogate-based Bayesian optimization, and gradient-based optimization methods.

\textbf{Evolutionary algorithm.}  
To the best of our knowledge, the only work that directly investigates the F4M optimization using evolutionary algorithms is the study by Liu et al.~\cite{liu2024many}, which proposed CluSO— a clustering-based swarm optimization method. This method explicitly addresses the challenge of handling many objectives with few solutions, and also introduced the DC-MaTS test suite to benchmark algorithmic performance. Our work follows this evolutionary approach and aims to further improve the search efficiency and solution quality by proposing a new evolutionary algorithm tailored for F4M optimization.

\textbf{Surrogate-based Bayesian optimization.}
More recently, Maus et al.~\cite{maus2025cover}
proposed Multi-Objective Coverage Bayesian Optimization
(MOCOBO), a Bayesian optimization framework for the
coverage optimization problem that is equivalent to the
F4M formulation in Eq.~\eqref{eq:sumofmin}.
MOCOBO builds Gaussian-process-based surrogate models
for each objective and designs an expected coverage
improvement (ECI) acquisition function to greedily
construct a covering set, combined with local trust-region
Bayesian optimization (TuRBO-style) to handle
high-dimensional continuous search spaces and structured
domains such as molecules and peptides. MOCOBO is designed to scale to large
numbers of function evaluations and to a large number
of objectives, which makes it conceptually applicable
to the F4M setting considered in this work as well.

\textbf{Gradient-based algorithms.}  
Another line of research focuses on solving the same type of problem—often referred to as sum-of-minimum (SoM) optimization—using gradient-based methods. Lin et al.~\cite{lin2024few} proposed Tchebycheff set scalarization, a novel scalarization technique designed to optimize a small solution set for many-objective problems. Similarly, Li et al.~\cite{li2024many} and Ding et al.~\cite{ding2024efficient} developed gradient-based frameworks targeting applications in machine learning, such as multi-task learning \cite{lin2019pareto} and mixture-of-prompt learning \cite{qin2021learning}.

Although these gradient-based methods provide elegant formulations and strong empirical results in continuous settings, they are primarily designed for applications where gradient information is accessible and reliable. In contrast, our work focuses on black-box optimization problems or combinatorial domains where gradients are not available or impractical to use \cite{bajaj2021black,korte2011combinatorial}. Therefore, we do not consider these gradient-based methods in our experimental studies.

\section{Proposed Algorithm}

This section describes the evolutionary algorithm developed to solve the F4M optimization problem. The algorithm is based on a $(\mu + 1)$ evolution strategy, integrates an archive mechanism to guide offspring generation and a removal strategy aligned with the F4M objective.

\subsection{Framework}
Inspired by the widely-adopted SMS-EMOA framework~\cite{beume2007sms}, our proposed algorithm employs a \((\mu+1)\)-evolution strategy designed explicitly for efficiently addressing the F4M optimization. The core principle of this strategy is to iteratively evolve a population of candidate solutions through offspring generation and selective solution removal, guided by the SoM objective function of F4M optimization defined in Eq.~(\ref{eq:sumofmin}). 

The detailed pseudo-code of our proposed algorithm, namely {Sum-of-Minimum (\underline{SoM}) driven \underline{E}volutionary \underline{M}ulti-objective \underline{O}ptimization \underline{A}lgorithm (SoM-EMOA)}, is provided in Algorithm~\ref{alg:MaCP_EMOA}.

\begin{algorithm}[h!]
\caption{{SoM-EMOA}}
\label{alg:MaCP_EMOA}
\begin{algorithmic}[1]
    \Statex\hspace*{-20pt} \textbf{Input:} Number of objectives \(m\), population size \(k\), maximum generations \(T\).
    \State Initialize solution set \(S\) with \(N\) random solutions where $N\gg k$.
    \State Evaluate objectives for all solutions in \(S\).
    \State \(A = \texttt{Archive\_initialization}(S)\).
    \State Randomly select $k$ solutions from $S$ to form the initial population $P$.
    \For{\(t=1\) to \(T\)}
        \State \(\boldsymbol{x}^{p_1},\boldsymbol{x}^{p_2}=\texttt{Mating\_selection}(P,A)\).
        \State \(\boldsymbol{x}^{\text{offspring}} = \texttt{Mutation}\left(\texttt{Crossover}(\boldsymbol{x}^{p_1}, \boldsymbol{x}^{p_2})\right)\).
        \State Evaluate objectives for \(\boldsymbol{x}^{\text{offspring}}\).
        \State \(A = \texttt{Archive\_update}(A,\boldsymbol{x}^{\text{offspring}})\).
        \State \(P' = P\cup \{\boldsymbol{x}^{\text{offspring}}\}\).
        \State Identify \(\boldsymbol{x}^*=\arg\min_{\boldsymbol{x}\in P'}G_{\text{ws}}\left(P'\setminus\{\boldsymbol{x}\}\right)\).
        \State Update population: \(P\leftarrow P'\setminus \{\boldsymbol{x}^*\}\).
    \EndFor
    \Statex\hspace*{-20pt} \textbf{Output:} The final population \(P\).
\end{algorithmic}
\end{algorithm}

The overall framework of our algorithm can be summarized in the following main steps:

\begin{enumerate}
    \item A population \( P \) consisting of \(k\) candidate solutions is randomly initialized within the feasible decision space.
    \item At each generation, exactly one offspring solution is generated using genetic operators and added to the current population, creating a temporary population of size \(k+1\).
    \item From this temporary population, a single solution is carefully removed to ensure that the remaining solutions optimally handle all objectives according to the objective function of F4M optimization in Eq.~(\ref{eq:sumofmin}).
    \item The process iterates until a predefined termination criterion (such as a maximum number of generations) is reached.
\end{enumerate}

Mathematically, let \(P=\{\boldsymbol{x}^{(1)}, \boldsymbol{x}^{(2)}, \dots, \boldsymbol{x}^{(k)}\}\) be the current population of candidate solutions. At each iteration, we generate one offspring solution \(\boldsymbol{x}^{\text{offspring}}\) and form a temporary population:
\[
P' = P \cup \{\boldsymbol{x}^{\text{offspring}}\}, \quad |P'| = k + 1.
\]

The next step is to remove a solution \(\boldsymbol{x}^*\) from \(P'\), so that the resulting population optimally addresses the F4M formulation, expressed as:
\[
\boldsymbol{x}^* = \arg\min_{\boldsymbol{x} \in P'} G_{\text{ws}}\left(P'\setminus\{\boldsymbol{x}\}\right).
\]

Below, we describe in detail how the offspring is generated and how the removal strategy is implemented.

\subsection{Archive Initialization}
{In traditional EMO algorithms, the offspring is generated by crossover and mutation based on the current population. However, this is not suitable for our algorithm since the population size $k$ is usually very small (e.g., 10). It is difficult to explore the search space based on a small population. In SoM-EMOA, we use an archive $A$ to assist the offspring generation. The archive $A$ serves as a memory of the best solutions encountered so far for each objective, which allows SoM-EMOA to generate offspring that explicitly target objectives that are currently under-represented in the population.}

{At the beginning of the search, a large temporary solution set $S$ of size $N \gg k$ is randomly generated.  
The archive is then initialized by selecting, for each objective $f_i(\boldsymbol{x})$, 
the solution in $S$ that achieves the best value on that objective:
\begin{equation}
    A = \left\{\arg\min_{\boldsymbol{x}\in S}f_1(\boldsymbol{x}),...,\arg\min_{\boldsymbol{x}\in S}f_m(\boldsymbol{x})\right\}.
\end{equation}}

{This ensures that each objective initially has at least one strong representative.  
This design is conceptually similar to the objective-wise elitist archives 
used in decomposition-based EMO algorithms~\cite{zhang2007moea} 
and in the initialization of multi-objective Bayesian optimization frameworks 
such as ParEGO~\cite{knowles2006parego}.  
The initialization provides a diverse yet targeted set of exemplars that 
guide subsequent search toward promising regions of the decision space.}

\subsection{Offspring Generation}


To efficiently generate high-quality offspring, we maintain a vector $\boldsymbol{u}\in\mathbb{R}^m$ which stores the minimum value for each objective in the archive $A$, i.e.,
\begin{equation}
    \boldsymbol{u} = \left(\min_{\boldsymbol{x}\in A}f_1(\boldsymbol{x}),...,\min_{\boldsymbol{x}\in A}f_m(\boldsymbol{x})\right),
\end{equation}
and a vector $\boldsymbol{v}\in\mathbb{R}^m$ which stores the minimum value for each objective in the current population $P$, i.e., 
\begin{equation}
    \boldsymbol{v} = \left(\min_{\boldsymbol{x}\in P} f_1(\boldsymbol{x}),... , \min_{\boldsymbol{x}\in P} f_m(\boldsymbol{x})\right).
\end{equation}

Therefore, we have $\boldsymbol{u}\leq \boldsymbol{v}$ and the vector $(\boldsymbol{v}-\boldsymbol{u})$ reflects the quality of the current population for addressing each objective. A large value (i.e., element) in $(\boldsymbol{v}-\boldsymbol{u})$ means that the current population cannot address the corresponding objective well. Thus, we can use this information to enhance the optimization process. Specifically, we utilize the information  $(\boldsymbol{v}-\boldsymbol{u})$ for mating selection, which is shown in Algorithm \ref{alg:mating-selection}.

In Algorithm \ref{alg:mating-selection}, we choose one solution $\boldsymbol{x}^{p_1}$ from the current population $P$ and one solution $\boldsymbol{x}^{p_2}$ from the archive $A$ as two parents. First, $\boldsymbol{x}^{p_1}$ is randomly selected from $P$. To select $\boldsymbol{x}^{p_2}$, we first calculate a probability distribution $\boldsymbol{p}$ over $m$ objectives as follows:
\begin{equation}
    \boldsymbol{p}=\frac{\boldsymbol{v}-\boldsymbol{u}}{\|\boldsymbol{v}-\boldsymbol{u}\|_1},
\end{equation}
where a large probability in $\boldsymbol{p}$ means that the corresponding objective is not well-addressed by the current population.  

Then we select $\boldsymbol{x}^{p_2}$ from the archive $A$ based on the probability distribution $\boldsymbol{p}$ so that we can enhance the optimization for the corresponding objective by $\boldsymbol{x}^{p_2}$. 

\begin{algorithm}[h!]
\caption{\(\boldsymbol{x}^{p_1},\boldsymbol{x}^{p_2}=\texttt{Mating\_selection}(P,A)\)}
\label{alg:mating-selection}
\begin{algorithmic}[1]
    \Statex\hspace*{-20pt} \textbf{Input:} Current population $P$, archive $A$, vector $\boldsymbol{u}\in\mathbb{R}^m$, and vector $\boldsymbol{v}\in\mathbb{R}^m$.
    \State Randomly selection one solution from $P$ as $\boldsymbol{x}^{p_1}$.
    \State Calculate the probability distribution $\boldsymbol{p}=\frac{\boldsymbol{v}-\boldsymbol{u}}{\|\boldsymbol{v}-\boldsymbol{u}\|_1}$.
    \State Select one solution from $A$ based on $\boldsymbol{p}$ as $\boldsymbol{x}^{p_2}$.
    \Statex\hspace*{-20pt} \textbf{Output:} Parent solutions $\boldsymbol{x}^{p_1}$ and $\boldsymbol{x}^{p_2}$.
\end{algorithmic}
\end{algorithm}

The offspring $\boldsymbol{x}^\text{offspring}$ is generated based on $\boldsymbol{x}^{p_1}$ and $\boldsymbol{x}^{p_2}$ using crossover and mutation. 

Our proposed mating selection approach enhances the efficiency of evolutionary search, directing exploration towards promising regions of the solution space.

\subsection{Archive Update}
{After each offspring is generated and evaluated, 
the archive is updated by comparing the offspring’s performance on every objective with the current archive members: 
\begin{equation}
    A = \left\{\arg\min_{\boldsymbol{x}\in A\cup\{\boldsymbol{x}^\text{offspring}\}}f_1(\boldsymbol{x}),...,\arg\min_{\boldsymbol{x}\in A\cup\{\boldsymbol{x}^\text{offspring}\}}f_m(\boldsymbol{x})\right\}.
\end{equation}}

{Thus, if the offspring improves any objective, it immediately replaces the previous best in $A$. This mechanism maintains per-objective elitism without storing the entire search history, 
and keeps the computational overhead linear in $m$.  
Conceptually, it parallels the ``reference-point elitism'' in RVEA~\cite{cheng2016reference} 
and the trust-region archive maintenance used in MOCOBO~\cite{maus2025cover}, 
but here it operates at the objective level instead of the region level.}


\subsection{Solution Removal Strategy}
Once the offspring is incorporated into the current population, a population reduction step is necessary. SoM-EMOA explicitly considers the objective function of F4M optimization in Eq.~(\ref{eq:sumofmin}) in the removal procedure. Specifically, we seek to remove one solution that leads to maximal improvement of the overall performance of F4M optimization, i.e., $\boldsymbol{x}^*=\arg\min_{\boldsymbol{x}\in P'}G_{\text{ws}}\left(P'\setminus\{\boldsymbol{x}\}\right)$.

However, calculating the removal criterion $G_{\text{ws}}$ from scratch for each solution $\boldsymbol{x} \in P'$ at each generation can be inefficient, especially for large \( m \). This is because for each solution $\boldsymbol{x} \in P'$, the time complexity of calculating $G_{\text{ws}}(P' \setminus \{\boldsymbol{x}\})$ from scratch is $\mathcal{O}(mk)$. Therefore, the total time complexity of identifying $\boldsymbol{x}^*$ is $\mathcal{O}(mk^2)$.

\begin{algorithm}[h!]
\caption{Identify \(\boldsymbol{x}^*=\arg\min_{\boldsymbol{x}\in P'}G_{\text{ws}}\left(P'\setminus\{\boldsymbol{x}\}\right)\)}
\label{alg:removal}
\begin{algorithmic}[1]
\Statex\hspace*{-20pt} \textbf{Input:} Current population $P$, offspring $\boldsymbol{x}^\text{offspring}$, $P' = P\cup\{\boldsymbol{x}^\text{offspring}\}$, vector $\boldsymbol{v}$.
    \State Calculate $G^\text{min} = \sum_{i=1}^m v_i$
    \State Set $\boldsymbol{x}^* = \boldsymbol{x}^\text{offspring}$ and $I=\emptyset$
    \For{$i \in \{1,...,m\}$}
    \If{$f_i(\boldsymbol{x}^\text{offspring})<=v_i$} 
        \State $v'_i = f_i(\boldsymbol{x}^\text{offspring})$
        \State $I = I\cup \{i\}$
    \EndIf
    \EndFor
    \If{$|I|==m$}
    \State Randomly select one solution from $P$ as $\boldsymbol{x}^*$
    \State Update $\boldsymbol{v} = \boldsymbol{v}'$
    \Else
    \For{$\boldsymbol{x}\in P$}
        \For{$i \in\{1,...,m\}\setminus I$}
        \If{$f_i(\boldsymbol{x})==v_i$}
        \State $v'_i = \min_{\boldsymbol{x}'\in P\setminus\{\boldsymbol{x}\}\cup \{\boldsymbol{x}^\text{offspring}\}} f_i(\boldsymbol{x}')$
        \Else
        \State $v'_i = v_i$
        \EndIf
        \EndFor
        \State Calculate $G_{\text{ws}}(P'\setminus\{\boldsymbol{x}\}) = \sum_{i=1}^m v'_i$
        \If{$G_{\text{ws}}(P'\setminus\{\boldsymbol{x}\})<G^\text{min}$}
        \State $G^\text{min}=G_{\text{ws}}(P'\setminus\{\boldsymbol{x}\})$
        \State Update $\boldsymbol{x}^* = \boldsymbol{x}$ and $\boldsymbol{v}^* = \boldsymbol{v}'$
        \EndIf
    \EndFor
    \State Update $\boldsymbol{v} = \boldsymbol{v}^*$
    \EndIf
    \Statex \hspace*{-17pt}\textbf{Output:} Solution $\boldsymbol{x}^*$ and vector $\boldsymbol{v}$. 
\end{algorithmic}
\end{algorithm}

To enhance the efficiency of identifying $\boldsymbol{x}^*$, we propose an efficient algorithm which is shown in Algorithm \ref{alg:removal}. The main idea in Algorithm \ref{alg:removal} is to avoid unnecessary comparisons in the calculation of $G_{\text{ws}}$.

In the {best case}, when the offspring \(\boldsymbol{x}^{\text{offspring}}\) improves all objectives (\(|I| = m\) in line 9), the algorithm executes only a constant-time update and random selection, resulting in a time complexity of \(\mathcal{O}(m)\). 
In the {worst case}, when multiple individuals attain the minimum value for some objectives, line~16 may re-scan almost the entire population for each individual and each objective, leading to a time complexity of \(\mathcal{O}(mk^2)\). 
In the {average case}, assuming objective values are drawn independently from continuous distributions, each minimum is unique with high probability, so the costly re-scan is expected to occur only once per objective. This yields an average-case time complexity of \(\mathcal{O}(mk)\).

\subsection{Distinctive Few-for-Many Characteristics}

{It is important to emphasize how the proposed SoM-EMOA
fundamentally differs from conventional multi-objective
or many-objective evolutionary algorithms.
Traditional EMO algorithms such as NSGA-II~\cite{deb2002fast},
MOEA/D~\cite{zhang2007moea}, or NSGA-III~\cite{deb2013evolutionary}
aim to approximate the entire Pareto front by maintaining
a large, diverse population that uniformly represents
trade-offs among all objectives.
In contrast, SoM-EMOA is designed under the \emph{few-for-many}
principle: its goal is not to represent every trade-off direction,
but to find a \emph{compact set of complementary solutions}
that collectively provide good coverage of all objectives.}

{From an algorithmic viewpoint, this conceptual difference
is reflected in three distinctive design elements:
\begin{enumerate}
    \item \textbf{Set-level objective evaluation.}
    Instead of evaluating individuals solely based on
    Pareto dominance or decomposition,
    SoM-EMOA explicitly optimizes the \emph{set-level coverage
    objective} $G_{\text{ws}}(X_k)$ defined in Eq.~\eqref{eq:sumofmin}.
    This objective measures how well the current population
    as a whole covers all $m$ objectives,
    ensuring that at least one solution performs well for each objective.
    Hence, the selection and removal operators directly operate
    on the coverage of the entire solution set,
    rather than individual fitness.
    \item \textbf{Complementarity-driven selection.}
    While conventional EMO algorithms reward diversity
    in the objective space (e.g., crowding distance, reference vectors),
    SoM-EMOA promotes diversity through \emph{complementarity}.
    A candidate is valuable if it improves the coverage
    of poorly addressed objectives.
    This is implemented through the probability-guided
    mating strategy (Algorithm~\ref{alg:mating-selection}),
    where offspring generation is biased toward objectives
    that are not well represented by the current population.
    \item \textbf{Compact population evolution.}
    Conventional EMO algorithms typically maintain large
    populations (hundreds or thousands of solutions)
    to approximate high-dimensional Pareto fronts.
    SoM-EMOA, however, deliberately operates with a small,
    fixed population size $k \ll m$,
    embodying the few-for-many setting.
    Each solution acts as a \emph{representative prototype}
    for a subset of objectives, analogous to cluster centroids
    in coverage optimization~\cite{liu2024many}.
\end{enumerate}}

{Through these mechanisms,
SoM-EMOA shifts the optimization focus from
approximating the entire Pareto set
to constructing a small, efficient covering set
that collectively represents the high-dimensional objective space.
This distinction captures the essential philosophy of
F4M optimization and differentiates SoM-EMOA
from conventional many-objective EMO frameworks.}

\section{A Benchmark Test Suite for F4M Optimization}

To evaluate the performance of F4M optimization algorithms, a suitable benchmark suite is essential. In this section, we first discuss the limitations of existing benchmarks and then introduce a new R2-based test suite tailored for F4M optimization, offering greater flexibility and challenge.
\subsection{Limitations of Existing Test Suites for F4M Optimization}
While numerous well-established benchmark suites exist for traditional multi-objective optimization, they are often not suitable for the F4M optimization.

First, standard multi-objective test suites such as DTLZ~\cite{deb2005scalable} and WFG~\cite{huband2006review} are inherently designed to evaluate algorithms on their ability to approximate the entire Pareto front. In many of these problems, it is often possible for a single solution to simultaneously perform well on a large number of objectives. For example, as pointed out by Ishibuchi et al. \cite{ishibuchi2016performance} and Liu et al.~\cite{liu2024many}, a single solution with $x_1=0$ can optimize the first $(m-1)$ objectives of DTLZ1. This behavior is unrealistic in real-world applications, where each objective often has a distinct optimum. Consequently, using such problems for evaluating F4M optimization algorithms may lead to misleading conclusions about their performance and generalizability.

To address the above issue, Liu et al.~\cite{liu2024many} proposed the Decoupled Many-objective Test Suite (DC-MaTS), which is specifically tailored for F4M optimization by ensuring that the optimal solution for each objective is independently located. This design makes it impossible for a single solution to optimize all objectives, thereby better reflecting the true nature of F4M optimization. However, despite its conceptual suitability, the DC-MaTS benchmark is relatively simple and exhibits low problem complexity. For example in DC-MaTS, there are no interactions between the decision variables and all the objective functions are unimodal. This simplicity limits its ability to differentiate the performance of more sophisticated F4M optimization solvers.

In light of these limitations, there is a clear need for more challenging and realistic test problems for F4M optimization. To this end, we propose a new R2-based test suite for F4M optimization in the following subsections. Our test suite is built upon the connection between the R2 indicator \cite{jaszkiewicz2024exact,schapermeier2024reinvestigating} and the objective of F4M optimization, and is highly flexible, allowing the transformation of any standard multi-objective optimization problem into a challenging F4M optimization instance with customizable complexity.

\subsection{Relation Between R2 Indicator and F4M Optimization}
The F4M optimization seeks to minimize the objective function $G_{\text{ws}}$ defined in Eq. \eqref{eq:sumofmin}. Interestingly, this formulation closely resembles the R2 indicator~\cite{jaszkiewicz2024exact,schapermeier2024reinvestigating}, a widely used performance metric in the evolutionary multi-objective optimization community. Specifically, the R2 indicator, when using Tchebycheff scalarization, is expressed as follows:
\begin{equation}\label{eq:R2}
    R_2(X, W, \boldsymbol{z}^*) = \frac{1}{|W|}\sum_{\boldsymbol{w} \in W}\min_{\boldsymbol{x} \in X}g_{\text{tch}}(\boldsymbol{x}|\boldsymbol{w},\boldsymbol{z}^*),
\end{equation}
where \( X \) is a set of candidate solutions, \( W \) is a set of weight vectors with each weight vector $\left\|\boldsymbol{w}\right\|_1=1$ and $\boldsymbol{w}\geq 0$, \( \boldsymbol{z}^* \) is the utopian point, and \( g_{\text{tch}}(\cdot) \) is the Tchebycheff scalarization function defined as:
\begin{equation}\label{eq:tch}
    g_{\text{tch}}(\boldsymbol{x}|\boldsymbol{w},\boldsymbol{z}^*) = \max_{1\leq i \leq m} \left\{ w_i |f_i(\boldsymbol{x}) - z^*_i| \right\}.
\end{equation}

Clearly, both the objective $G_{\text{ws}}(X_k)$ of F4M optimization in Eq.~(\ref{eq:sumofmin}) and the R2 indicator in Eq.~(\ref{eq:R2}) involve the sum-of-minimum form. The similarity indicates that the R2 indicator can be adapted to F4M optimization to develop systematic test suites for evaluating F4M optimization algorithms.

\subsection{Proposed F4M Optimization Benchmark Test Suite}

Motivated by the close relationship between the F4M optimization formulation and the R2 indicator, we propose a benchmark test suite specifically tailored for F4M optimization based on the R2 indicator formulation.

The proposed test suite is constructed using the following steps:

    \textbf{Step 1:} Generate a set of weight vectors \( W = \{\boldsymbol{w}^{(1)}, \boldsymbol{w}^{(2)}, \dots, \boldsymbol{w}^{(m)}\} \) using a standard method such as simplex lattice design~\cite{das1998normal} or uniform distribution~\cite{saini2025efficient}.
    
    \textbf{Step 2:} Define the objective functions for the F4M optimization problem using the Tchebycheff scalarization associated with each weight vector. Specifically, given an original $q$-objective optimization problem (e.g., DTLZ and WFG):
    \[
    \min_{\boldsymbol{x} \in \mathcal{X}} \boldsymbol{f}(\boldsymbol{x}) = (f_1(\boldsymbol{x}), f_2(\boldsymbol{x}), \dots, f_q(\boldsymbol{x})),
    \]
    we define each new objective function \(F_i(\boldsymbol{x})\) of the F4M optimization as:
    \begin{equation}\label{eq:MaCP_test_objective}
    \begin{aligned}
        F_i(\boldsymbol{x}) &= g_{\text{tch}}(\boldsymbol{x}|\boldsymbol{w}^{(i)}, \boldsymbol{z}^*)\\
        &= \max_{1\leq j\leq q}\{ w^{(i)}_j |f_j(\boldsymbol{x})-z^*_j|\}, i=1,\dots,m,
    \end{aligned}
    \end{equation}
    where \( \boldsymbol{z}^* \) is the utopian point in the original \( q \)-objective space.
    
    \textbf{Step 3:} Formulate the resulting F4M optimization problem as:
    \begin{equation}\label{eq:MaCP_test}
        \min_{X_k=\{\boldsymbol{x}^{(1)},\dots,\boldsymbol{x}^{(k)}\}\subseteq \mathcal{X}} \sum_{i=1}^{m}\min_{1\leq j \leq k}F_i(\boldsymbol{x}^{(j)}).
    \end{equation}

By following this procedure, any existing multi-objective optimization problem can be systematically transformed into a F4M optimization problem, greatly enhancing the flexibility and versatility of our proposed test suite. The flexibility of our test suite allows the systematic investigation of algorithmic behavior across a wide range of characteristics. This facilitates comprehensive evaluations and promotes rigorous benchmarking of F4M optimization algorithms.



\subsection{Example: Transforming a 2-Objective DTLZ2 Problem into an F4M Instance}

{To illustrate how a standard multi-objective optimization problem (MOP) can be transformed into an 
F4M instance under the proposed benchmark framework,
consider the 2-objective DTLZ2 problem~\cite{deb2005scalable}.}

{The 2-objective DTLZ2 problem is defined as
\begin{equation}
    \min_{\boldsymbol{x} \in [0,1]^d}
    \begin{cases}
        f_1(\boldsymbol{x}) = (1 + g(x_M)) \cos\!\left(\tfrac{\pi}{2} x_1 \right), \\[2mm]
        f_2(\boldsymbol{x}) = (1 + g(x_M)) \sin\!\left(\tfrac{\pi}{2} x_1 \right),
    \end{cases}
\end{equation}
where $g(x_M)$ depends on the last $(d-1)$ variables.
For all Pareto-optimal solutions $\boldsymbol{x}$ with $g(x_M)=0$, 
the Pareto front satisfies
\begin{equation}
    f_1^2 + f_2^2 = 1, \quad f_1,f_2 \ge 0,
\end{equation}
that is, a quarter-circle arc in the first quadrant of the objective space.}

{In conventional multi-objective optimization,
algorithms such as NSGA-II or MOEA/D aim to approximate this continuous Pareto front 
by producing a large set of nondominated points distributed along the arc. Under the F4M paradigm, we replace the goal of approximating the entire front 
with the goal of selecting a small number of representative solutions that collectively cover 
a much larger number of scalarized objectives.}

\begin{figure}[htbp]
    \centering
\includegraphics[width=0.7\linewidth]{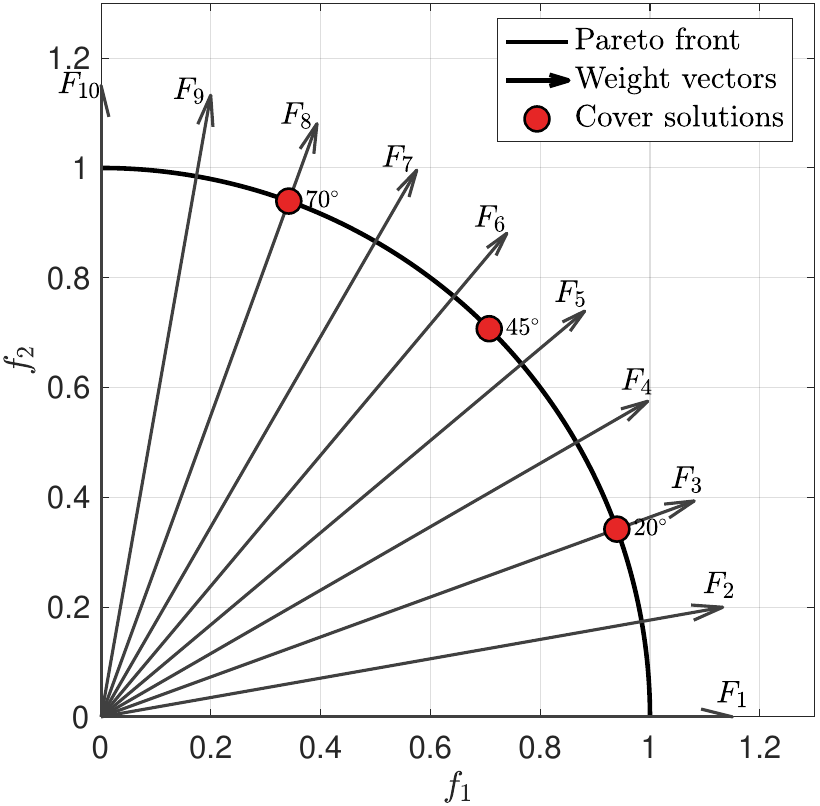} 
\caption{{
Illustration of the F4M formulation on the 2-objective DTLZ2 problem. 
The quarter-circle curve represents the Pareto front of DTLZ2. 
Ten directional scalarized objectives $F_1,\ldots,F_{10}$ are shown as 
arrows originating from the utopian point and pointing toward uniformly 
distributed directions on the front. 
A small set of three representative solutions located near 
$20^\circ$, $45^\circ$, and $70^\circ$ collectively covers these 
directional objectives, demonstrating how F4M approximates a continuous 
Pareto front using a compact solution set.
}}
    \label{fig:example}
\end{figure}

{To do so, we generate $m$ uniformly distributed weight vectors. Each weight vector $\boldsymbol{w}^{(i)}$ defines one scalarized objective function $F_i(\boldsymbol{x})$ as in Eq. \eqref{eq:MaCP_test_objective}. Here, each $F_i(\boldsymbol{x})$ corresponds to a weight vector $\boldsymbol{w}^{(i)}$. For example, if we set $m=10$ equally spaced weight vectors as illustrated in Fig. \ref{fig:example},
we obtain 10 coverage objectives $F_1,F_2,...,F_{10}$ representing 10 evenly distributed directions 
on the quarter-circle. The F4M objective in Eq.~\eqref{eq:MaCP_test} 
seeks a small number $k$ of points on or near the arc 
such that every direction $\boldsymbol{w}^{(i)}$ has at least one solution 
that performs well along it.
When $k=3$, for instance, the algorithm is encouraged to find 
three solutions positioned roughly at $20^\circ$, $45^\circ$, and $70^\circ$ on the arc as shown in Fig. \ref{fig:example}, 
so that their union collectively provides good coverage of the 10 scalarized objectives.
This setting demonstrates how a continuous Pareto front can be 
approximated by a compact, complementary solution set under the F4M formulation. }

\section{Experiments}

In this section, we present a comprehensive empirical evaluation of our proposed algorithm. Our main goal is to demonstrate the effectiveness and efficiency of the proposed algorithm for F4M optimization, compared with state-of-the-art evolutionary algorithms.

\subsection{Experimental Settings}

To ensure a fair and consistent comparison, we closely follow the experimental setup described in the recent study on F4M optimization by Liu et al.~\cite{liu2024many}, where the CluSO algorithm was introduced. In particular, we adopt the same parameter configurations and performance metric. All experiments are performed on PlatEMO \cite{tian2017platemo}.

\subsubsection{Benchmark Problems} 
We use existing synthetic benchmark problems and also the proposed R2-based F4M optimization test suite described in Section~IV. For existing benchmark problems, we choose DC-MaTS problems proposed in \cite{liu2024many} and the noisy mixed linear regression (NMLR) problem considered in \cite{lin2024few}. For the proposed test suite, the widely used DTLZ \cite{deb2005scalable} and WFG \cite{huband2006review} problems are transformed into F4M optimization instances (denoted as F4M-DTLZ and F4M-WFG) using scalarization functions associated with a set of uniformly distributed weight vectors. The number of objectives for the original DTLZ and WFG problems is set to $q=3$. The number of decision variables for F4M-DTLZ and F4M-WFG are set to their default values based on $q$. The number of objectives for all F4M optimization problems is set to $m\in \{25, 50, 75, 100\}$. 

{In addition, we also consider two real-world many-objective optimization problems, DDMOP1 and DDMOP4, proposed in \cite{he2020repository}. DDMOP1 is a vehicle performance optimization problem known as the car cab design problem, which involves 11 decision variables and 9 objectives. The decision variables describe structural and dimensional parameters of the car body (e.g., thickness of the B-pillar, floor side inner, and door beam), while the objectives characterize key performance aspects such as weight, fuel economy, acceleration time, noise level, and cabin roominess. DDMOP4, on the other hand, is an electronic circuit design problem concerning an LTLCL switching ripple suppressor with nine resonant branches. It contains 13 decision variables and 10 objectives, where the design variables correspond to component parameters such as capacitances, inductances, and resistances, and the objectives aim to minimize the total inductor cost and the harmonic amplitudes at multiple resonant frequencies. Both problems are highly complex, black-box, and computationally expensive.}

\subsubsection{Compared Algorithms}
We consider the following algorithms for comparison in our experiments.
\begin{itemize}
    \item {CluSO}~\cite{liu2024many}: The clustering-based swarm optimizer introduced as the first dedicated evolutionary algorithm for F4M optimization.
    \item {{MOCOBO}\footnote{https://github.com/nataliemaus/mocobo}~\cite{maus2025cover}: A Bayesian optimization-based method that constructs multiple trust regions and trains Gaussian process surrogates to iteratively improve a small set of covering solutions. }
    \item NSGA-II \cite{deb2002fast}, PREA \cite{yuan2020investigating}, {NRV-MOEA}~\cite{hua2024adaptive}, {MOEA/D-UR}~\cite{de2022decomposition}, and HEA \cite{liu2023many}: State-of-the-art EMO algorithms included for reference, although they are not specifically designed for F4M optimization. These algorithms aim to approximate the entire Pareto front of the $m$-objective problems rather than optimize a small covering set.
\end{itemize}

It is important to emphasize that among the compared algorithms, only CluSO, MOCOBO and SoM-EMOA are explicitly designed to solve F4M optimization problems and are aware of the specific objective function \( G_{\text{ws}}(X_k) \). In contrast, the other EMO algorithms (NSGA-II, PREA, NRV-MOEA, MOEA/D-UR, and HEA) are not tailored for the F4M setting and do not incorporate any mechanisms to directly optimize \( G_{\text{ws}}(X_k) \). As a result, they lack the ability to intentionally construct a small, complementary solution set that jointly covers all objectives.

\subsubsection{Population and Termination} For the synthetic test problems, the population size is set to \(600\), and the maximum number of function evaluations is set to \(60,000\). {For the two real-world problems, the population size is set to 100, and the maximum number of function evaluations is set to \(1000\) and 10000.} The size of the final output set is set as \(k \in \{ 5, 10\}\) to emphasize the goal of F4M optimization: handling many objectives with a few solutions. For those traditional EMO algorithms, their final output is the final population. For fair comparison, we use a greedy subset selection algorithm described in Algorithm \ref{alg:selection} to choose $k$ solutions from the final population as the final output of those traditional EMO algorithms.

\begin{algorithm}[h!]
\caption{Greedy F4M Subset Selection}
\label{alg:selection}
\begin{algorithmic}[1]
\Statex\hspace*{-20pt} \textbf{Input:} Final population $P$.
\State Initialize $S=\emptyset$.
\For{$i\in\{1,...,k\}$}
\State Identify $\boldsymbol{x}^* = \arg\min_{\boldsymbol{x}\in P} G_{\text{ws}}(\{\boldsymbol{x}\}\cup S)$
\State Update $S = S\cup\{\boldsymbol{x}^*\}$ and $P = P\setminus\{\boldsymbol{x}^*\}$
\EndFor
\Statex\hspace*{-20pt} \textbf{Output:} A subset $S\subset P$ with $|S|=k$.
\end{algorithmic}
\end{algorithm}

\begin{table*}[htbp]
\renewcommand{\arraystretch}{1.2}
\centering
\caption{The mean values of $G_{\text{ws}}(X_k)$ obtained by SoM-EMOA and all baselines on existing benchmark test problems. The size of the final solution set $k$ is 5. The `$+$', `$-$' and `$\approx$' indicate that the compared algorithm is `significantly better than', `significantly worse than' and `statistically similar to' SoM-EMOA, respectively.}
\begin{tabular}{cccccccccc}
\toprule
Problem&$m$&$d$&NSGA-II&PREA&NRV-MOEA&MOEA/D-UR&HEA&CluSO&SoM-EMOA\\
\midrule
\multirow{4}{*}{DC-MaTS1}&25&25&-1.6042e+1 $-$&-1.6020e+1 $-$&-1.0927e+1 $-$&-1.4026e+1 $-$&-4.4803e+0 $-$&-1.7131e+1 $-$&\hl{-2.0761e+1}\\
&50&50&-2.8409e+1 $-$&-3.0261e+1 $-$&-2.1476e+1 $-$&-2.5645e+1 $-$&-2.2579e+0 $-$&-3.3120e+1 $-$&\hl{-3.9372e+1}\\
&75&75&-3.9786e+1 $-$&-3.8102e+1 $-$&-3.0181e+1 $-$&-3.6454e+1 $-$&-2.6148e+1 $-$&-4.8496e+1 $-$&\hl{-5.4906e+1}\\
&100&100&-5.0166e+1 $-$&-4.3673e+1 $-$&-3.9080e+1 $-$&-4.6837e+1 $-$&-3.5476e+1 $-$&-6.3067e+1 $-$&\hl{-6.9271e+1}\\
\hline
\multirow{4}{*}{DC-MaTS2}&25&25&-1.6050e+1 $-$&-1.6233e+1 $-$&-1.2472e+1 $-$&-1.5493e+1 $-$&-7.0205e+0 $-$&-1.7522e+1 $-$&\hl{-2.0725e+1}\\
&50&50&-2.9340e+1 $-$&-3.1975e+1 $-$&-2.5706e+1 $-$&-2.6801e+1 $-$&-1.8118e+1 $-$&-3.1545e+1 $-$&\hl{-4.0098e+1}\\
&75&75&-4.1662e+1 $-$&-3.9802e+1 $-$&-3.5373e+1 $-$&-3.7344e+1 $-$&-3.1877e+1 $-$&-4.5411e+1 $-$&\hl{-5.8179e+1}\\
&100&100&-5.2844e+1 $-$&-4.5429e+1 $-$&-4.5627e+1 $-$&-4.7871e+1 $-$&-4.2089e+1 $-$&-5.8587e+1 $-$&\hl{-7.4708e+1}\\
\hline
\multirow{4}{*}{DC-MaTS3}&25&25&-1.5735e+1 $-$&-1.4719e+1 $-$&-1.0936e+1 $-$&-1.2712e+1 $-$&-3.8945e+0 $-$&-1.6430e+1 $-$&\hl{-2.0761e+1}\\
&50&50&-2.7522e+1 $-$&-2.7469e+1 $-$&-1.7230e+1 $-$&-1.8153e+1 $-$&-3.0204e+0 $-$&-3.0680e+1 $-$&\hl{-3.8744e+1}\\
&75&75&-3.7624e+1 $-$&-3.6953e+1 $-$&-2.2956e+1 $-$&-2.7893e+1 $-$&-1.7042e+1 $-$&-4.3976e+1 $-$&\hl{-5.0979e+1}\\
&100&100&-4.6606e+1 $-$&-4.4198e+1 $-$&-2.9136e+1 $-$&-3.6269e+1 $-$&-2.5452e+1 $-$&-5.9525e+1 $-$&\hl{-6.1623e+1}\\
\hline
\multirow{4}{*}{DC-MaTS4}&25&25&-1.6315e+1 $-$&-1.7080e+1 $-$&-1.1722e+1 $-$&-1.3852e+1 $-$&-5.8713e+0 $-$&-1.6733e+1 $-$&\hl{-2.0758e+1}\\
&50&50&-2.8999e+1 $-$&-2.9934e+1 $-$&-2.0039e+1 $-$&-2.6205e+1 $-$&-2.0596e+0 $-$&-3.0800e+1 $-$&\hl{-4.0054e+1}\\
&75&75&-4.0553e+1 $-$&-3.5309e+1 $-$&-2.8575e+1 $-$&-3.9063e+1 $-$&-2.6617e+1 $-$&-4.4203e+1 $-$&\hl{-5.4265e+1}\\
&100&100&-5.1148e+1 $-$&-3.9836e+1 $-$&-3.6679e+1 $-$&-5.0318e+1 $-$&-3.3677e+1 $-$&-6.2014e+1 $-$&\hl{-6.6698e+1}\\
\hline
\multirow{4}{*}{NMLR}&25&10&1.8637e+0 $-$&9.4321e-1 $-$&1.2388e+0 $-$&1.1468e+0 $-$&1.2590e+0 $-$&6.0168e-1 $-$&\hl{2.0673e-1}\\
&50&10&5.8519e+0 $-$&3.6759e+0 $-$&4.1019e+0 $-$&4.2455e+0 $-$&6.1741e+0 $-$&2.0714e+0 $-$&\hl{7.9513e-1}\\
&75&10&1.3432e+1 $-$&8.9401e+0 $-$&1.0944e+1 $-$&1.0582e+1 $-$&1.4440e+1 $-$&5.1172e+0 $-$&\hl{2.1562e+0}\\
&100&10&1.7722e+1 $-$&1.1892e+1 $-$&1.4738e+1 $-$&1.7212e+1 $-$&1.9174e+1 $-$&1.1357e+1 $-$&\hl{3.7980e+0}\\
\hline
\multicolumn{3}{c}{$+/-/\approx$}&0/20/0&0/20/0&0/20/0&0/20/0&0/20/0&0/20/0&-\\
\bottomrule
\label{table:comapre}
\end{tabular}
\end{table*}

\subsubsection{Variation Operators} All algorithms employ the Simulated Binary Crossover (SBX) \cite{pan2021adaptive} with distribution index \(\eta_c = 20\) and the Polynomial Mutation \cite{carles2023self} with index \(\eta_m = 20\), consistent with the settings in~\cite{liu2024many}. The crossover probability is set to \(p_c = 1.0\), and the mutation probability is \(p_m = 1/d\) where $d$ is the number of decision variables.

\subsubsection{Performance Metric} Following~\cite{liu2024many}, the performance of each algorithm is evaluated using the SoM function \( G_{\text{ws}}(X_k) \) defined in Eq. \eqref{eq:sumofmin}. A smaller \( G_{\text{ws}}(X_k) \) value indicates better overall objective coverage.

Each algorithm is independently executed 30 times, and the mean and standard deviation of the performance metric are reported to ensure statistical reliability. All the results are tested based on the Wilcoxon rank-sum test at a 0.05 significance level to verify the statistical significance of performance differences.

\subsection{Results on Synthetic Problems}

\begin{figure}[!htb]
    \centering
    \subfigure[DC-MaTS1]{ 
\includegraphics[width=0.48\linewidth]{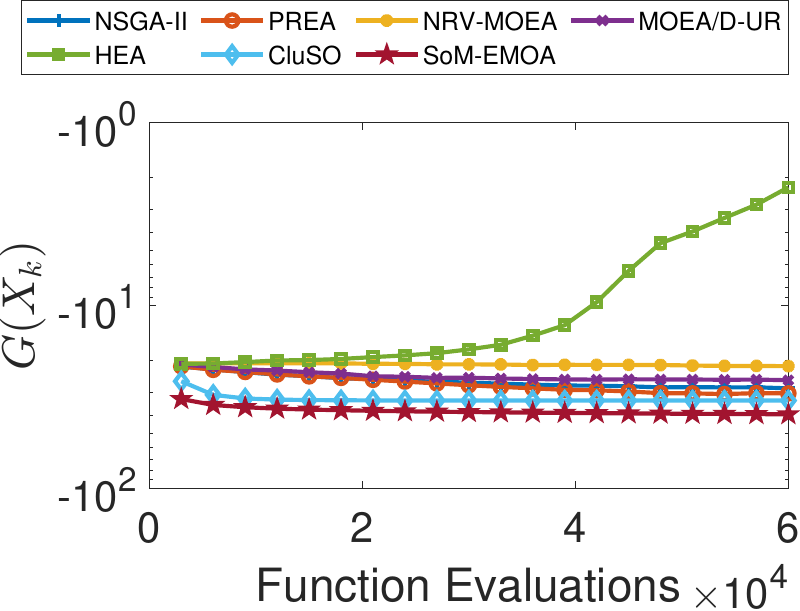} }
\hspace{-8pt}\subfigure[DC-MaTS2]{ 
\includegraphics[width=0.48\linewidth]{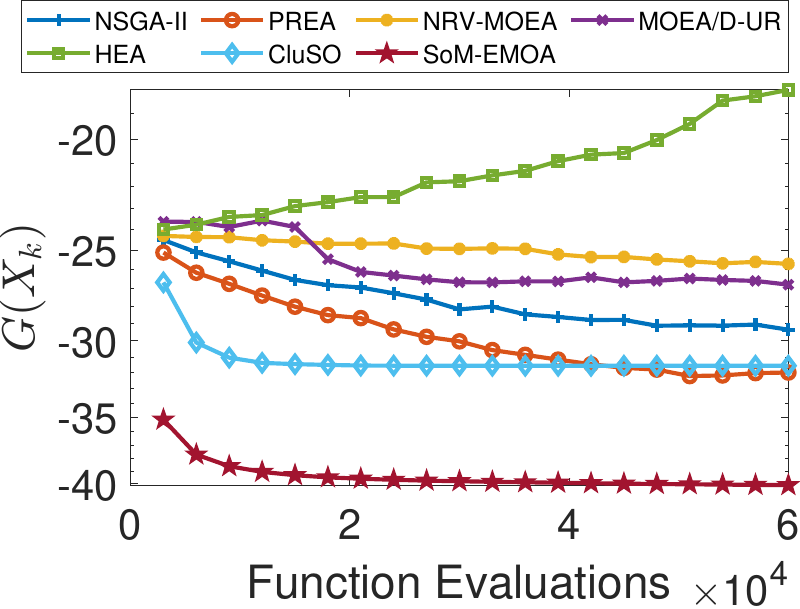} }
 \hspace{-8pt}\subfigure[DC-MaTS3]{ 
\includegraphics[width=0.48\linewidth]{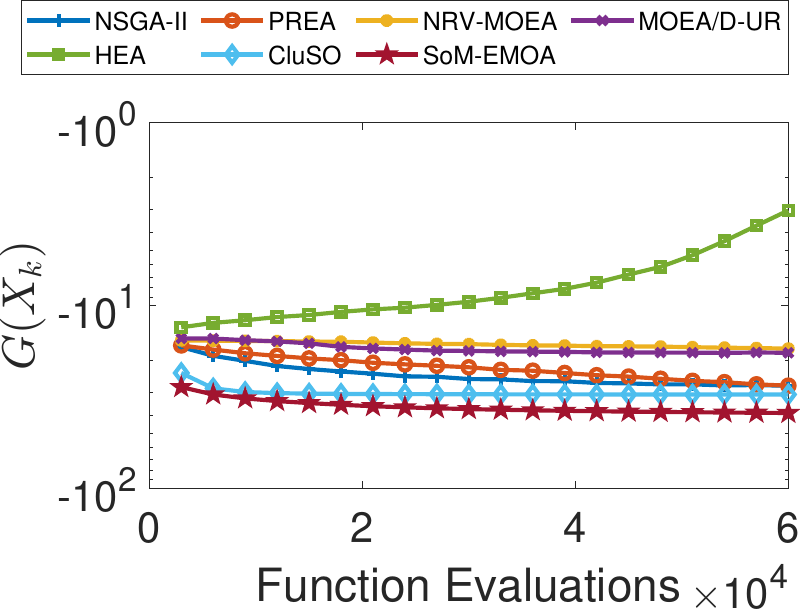} }
 \hspace{-8pt}\subfigure[DC-MaTS4]{ 
\includegraphics[width=0.48\linewidth]{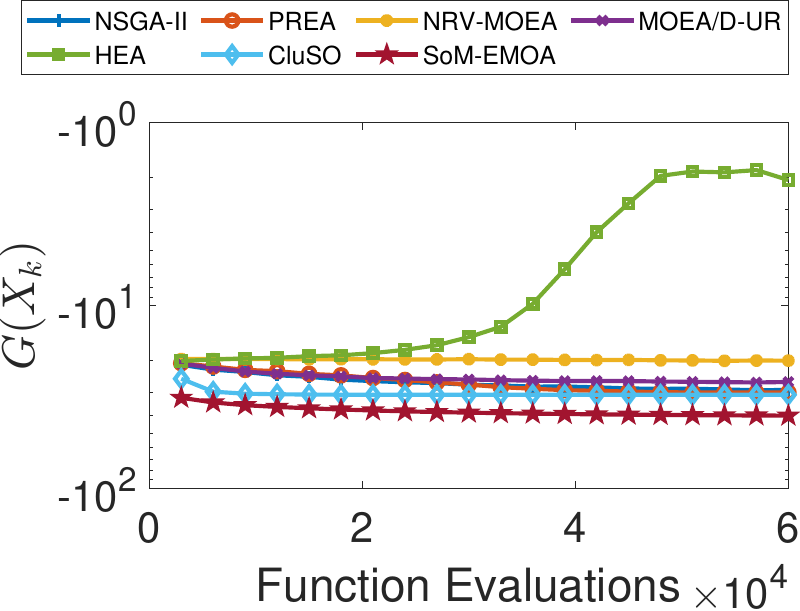} }
\subfigure[NMLR]{ 
\includegraphics[width=0.48\linewidth]{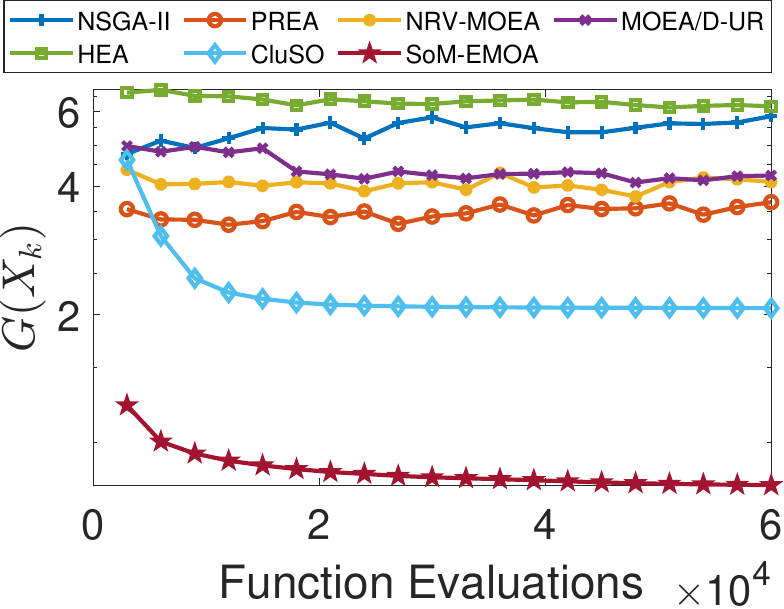}   }            
    \caption{{The convergence curve of different algorithms on 5 existing F4M optimization benchmark problems. The number of objectives $m=50$ and the size of the solution set $k=5$.}}
    \label{fig:compare}
\end{figure}

\begin{table*}[!htb]
\renewcommand{\arraystretch}{1.2}
\centering
\caption{The mean values of $G_{\text{ws}}(X_k)$ obtained by SoM-EMOA and all baselines on the proposed benchmark test problems. The size of the final solution set $k$ is 5. The `$+$', `$-$' and `$\approx$' indicate that the compared algorithm is `significantly better than', `significantly worse than' and `statistically similar to' SoM-EMOA, respectively.}
\begin{tabular}{cccccccccc}
\toprule
Problem&$m$&$d$&NSGA-II&PREA&NRV-MOEA&MOEA/D-UR&HEA&CluSO&SoM-EMOA\\
\midrule
\multirow{4}{*}{F4M-DTLZ1}&25&7&1.0045e+0 $-$&1.0143e+0 $-$&1.0111e+0 $-$&1.0249e+0 $-$&1.0160e+0 $-$&1.3197e+2 $-$&\hl{9.4407e-1}\\
&50&7&2.1148e+0 $-$&2.1359e+0 $-$&2.1264e+0 $-$&2.3078e+0 $-$&2.1348e+0 $-$&2.6285e+2 $-$&\hl{2.0601e+0}\\
&75&7&3.2388e+0 $-$&3.2555e+0 $-$&3.2596e+0 $-$&3.5789e+0 $-$&3.2910e+0 $-$&3.8055e+2 $-$&\hl{3.1594e+0}\\
&100&7&4.3598e+0 $-$&4.3772e+0 $-$&4.3861e+0 $-$&4.9325e+0 $-$&4.4170e+0 $-$&5.2429e+2 $-$&\hl{4.2456e+0}\\
\hline
\multirow{4}{*}{F4M-DTLZ2}&25&12&2.3841e+0 $-$&2.3819e+0 $-$&2.4047e+0 $-$&2.4079e+0 $-$&2.4027e+0 $-$&4.2317e+0 $-$&\hl{2.3770e+0}\\
&50&12&5.0278e+0 $-$&5.0280e+0 $-$&5.0814e+0 $-$&5.0631e+0 $-$&5.0565e+0 $-$&9.0217e+0 $-$&\hl{5.0055e+0}\\
&75&12&7.6872e+0 $-$&7.6869e+0 $-$&7.7445e+0 $-$&7.9240e+0 $-$&7.7149e+0 $-$&1.4346e+1 $-$&\hl{7.6692e+0}\\
&100&12&1.0354e+1 $-$&1.0347e+1 $-$&1.0454e+1 $-$&1.0665e+1 $-$&1.0371e+1 $-$&1.8970e+1 $-$&\hl{1.0316e+1}\\
\hline
\multirow{4}{*}{F4M-DTLZ3}&25&12&2.9496e+0 $-$&1.0418e+1 $-$&4.8099e+0 $-$&2.5187e+0 $-$&2.5187e+0 $-$&1.0286e+3 $-$&\hl{2.3900e+0}\\
&50&12&5.7855e+0 $-$&1.9300e+1 $-$&1.0721e+1 $-$&6.5438e+0 $-$&5.0825e+0 $-$&2.2005e+3 $-$&\hl{5.0234e+0}\\
&75&12&8.8362e+0 $-$&2.7879e+1 $-$&1.5911e+1 $-$&1.2696e+1 $-$&7.7517e+0 $-$&3.2520e+3 $-$&\hl{7.6933e+0}\\
&100&12&1.2081e+1 $-$&3.7592e+1 $-$&2.0053e+1 $-$&1.8583e+1 $-$&1.0424e+1 $-$&4.5370e+3 $-$&\hl{1.0352e+1}\\
\hline
\multirow{4}{*}{F4M-DTLZ4}&25&12&2.3846e+0 $-$&2.3818e+0 $-$&2.4054e+0 $-$&2.3944e+0 $-$&2.4856e+0 $-$&3.4705e+0 $-$&\hl{2.3773e+0}\\
&50&12&5.0281e+0 $-$&5.0265e+0 $-$&5.0894e+0 $-$&5.0563e+0 $-$&6.2525e+0 $-$&7.7132e+0 $-$&\hl{5.0043e+0}\\
&75&12&7.6877e+0 $-$&7.6831e+0 $-$&7.7748e+0 $-$&7.9817e+0 $-$&9.5849e+0 $-$&1.2368e+1 $-$&\hl{7.6689e+0}\\
&100&12&1.0353e+1 $+$&\hl{1.0345e+1 $+$}&1.0463e+1 $+$&1.0924e+1 $-$&1.1163e+1 $-$&1.5875e+1 $-$&1.0476e+1\\
\hline
\multirow{4}{*}{F4M-WFG1}&25&12&5.1876e+0 $-$&6.1981e+0 $-$&5.5615e+0 $-$&4.5628e+0 $-$&4.5723e+0 $-$&1.8780e+1 $-$&\hl{4.3270e+0}\\
&50&12&1.0594e+1 $-$&1.2581e+1 $-$&1.1296e+1 $-$&9.4794e+0 $-$&9.5165e+0 $-$&3.6270e+1 $-$&\hl{9.0636e+0}\\
&75&12&1.5813e+1 $-$&1.8845e+1 $-$&1.7101e+1 $-$&1.4907e+1 $-$&1.4477e+1 $-$&5.5646e+1 $-$&\hl{1.3769e+1}\\
&100&12&2.1259e+1 $-$&2.5031e+1 $-$&2.2606e+1 $-$&2.0061e+1 $-$&1.9507e+1 $-$&7.3400e+1 $-$&\hl{1.8475e+1}\\
\hline
\multirow{4}{*}{F4M-WFG2}&25&12&4.6982e+0 $-$&4.6933e+0 $-$&4.7613e+0 $-$&4.6833e+0 $-$&4.6814e+0 $-$&7.9513e+0 $-$&\hl{4.5091e+0}\\
&50&12&9.8864e+0 $-$&9.9091e+0 $-$&1.0024e+1 $-$&9.8948e+0 $-$&9.9535e+0 $-$&1.7959e+1 $-$&\hl{9.6015e+0}\\
&75&12&1.4989e+1 $-$&1.5046e+1 $-$&1.5239e+1 $-$&1.5055e+1 $-$&1.5089e+1 $-$&2.7461e+1 $-$&\hl{1.4602e+1}\\
&100&12&2.0098e+1 $-$&2.0171e+1 $-$&2.0405e+1 $-$&2.0175e+1 $-$&2.0189e+1 $-$&3.6695e+1 $-$&\hl{1.9637e+1}\\
\hline
\multirow{4}{*}{F4M-WFG3}&25&12&1.0055e+1 $-$&1.0139e+1 $-$&1.0334e+1 $-$&1.0278e+1 $-$&1.0376e+1 $-$&1.2404e+1 $-$&\hl{9.8869e+0}\\
&50&12&2.1681e+1 $-$&2.1863e+1 $-$&2.2180e+1 $-$&2.2644e+1 $-$&2.2201e+1 $-$&2.6348e+1 $-$&\hl{2.1094e+1}\\
&75&12&3.2966e+1 $-$&3.3307e+1 $-$&3.3734e+1 $-$&3.5368e+1 $-$&3.3751e+1 $-$&4.0618e+1 $-$&\hl{3.1778e+1}\\
&100&12&4.4319e+1 $-$&4.4562e+1 $-$&4.5289e+1 $-$&4.7456e+1 $-$&4.5413e+1 $-$&5.2939e+1 $-$&\hl{4.2688e+1}\\
\hline
\multirow{4}{*}{F4M-WFG4}&25&12&7.7557e+0 $-$&7.7475e+0 $-$&7.8386e+0 $-$&7.8083e+0 $-$&7.7003e+0 $-$&1.0602e+1 $-$&\hl{7.6640e+0}\\
&50&12&1.6446e+1 $-$&1.6409e+1 $-$&1.6609e+1 $-$&1.6540e+1 $-$&1.6638e+1 $-$&2.3528e+1 $-$&\hl{1.6294e+1}\\
&75&12&2.5069e+1 $-$&2.5042e+1 $-$&2.5318e+1 $-$&2.5483e+1 $-$&2.5245e+1 $-$&3.7768e+1 $-$&\hl{2.4913e+1}\\
&100&12&3.3648e+1 $-$&3.3633e+1 $-$&3.4006e+1 $-$&3.4392e+1 $-$&3.3972e+1 $-$&5.1221e+1 $-$&\hl{3.3445e+1}\\
\hline
\multicolumn{3}{c}{$+/-/\approx$}&1/31/0&1/31/0&1/31/0&0/32/0&0/32/0&0/32/0&-\\
\bottomrule
\label{table:comapre1}
\end{tabular}
\end{table*}

Table~\ref{table:comapre} summarizes the mean values of $G_{\text{ws}}(X_k)$ obtained by SoM-EMOA and six baseline algorithms across 5 existing benchmark problems for the case of $k=5$ (The results for $k=10$  are provided in the supplementary material). As shown in Table~\ref{table:comapre}, SoM-EMOA significantly outperforms all compared algorithms on every test problem. Specifically, it achieves the lowest $G_{\text{ws}}(X_k)$ value on all 5 problems, with statistically significant improvement over each baseline. Notably, even CluSO—the most competitive existing F4M solver—fails to outperform SoM-EMOA in any case.

The convergence behaviors of all algorithms on existing benchmark problems with $m=50$ and $k=5$ are illustrated in Fig.~\ref{fig:compare}. Across all instances, SoM-EMOA exhibits a faster and more stable reduction in $G_{\text{ws}}(X_k)$ compared to the baselines. This highlights its efficiency in discovering high-quality solutions during the evolutionary process.

\begin{figure*}[htbp]
    \centering
 \hspace{-8pt}\subfigure[F4M-DTLZ1]{ 
\includegraphics[width=0.24\linewidth]{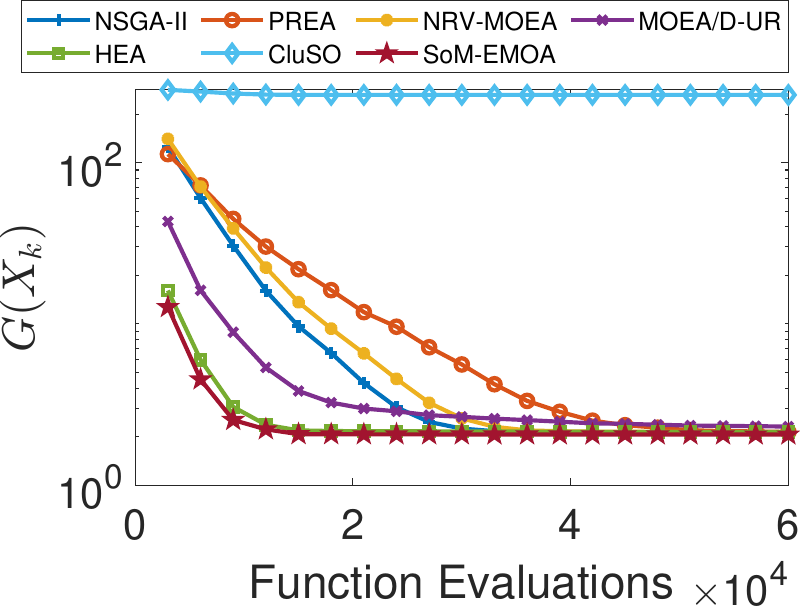} }
 \hspace{-8pt}\subfigure[F4M-DTLZ2]{ 
\includegraphics[width=0.24\linewidth]{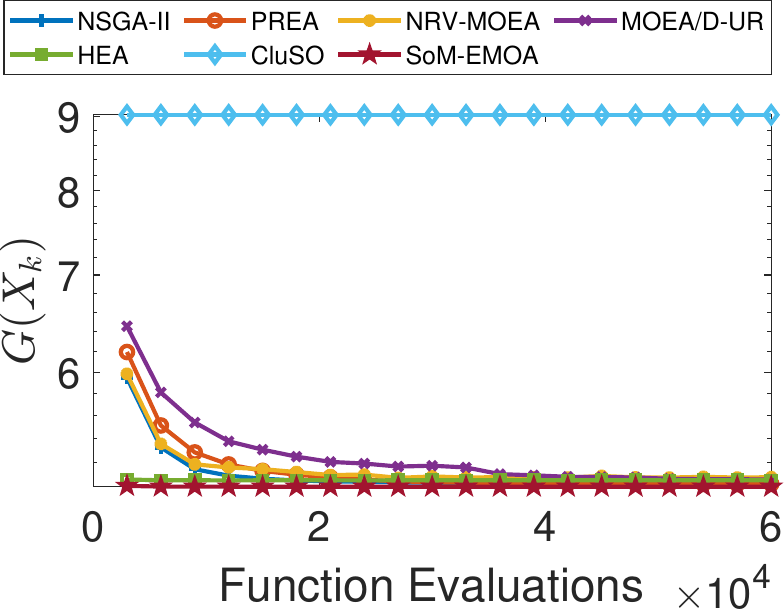} }
 \hspace{-8pt}\subfigure[F4M-DTLZ3]{ 
\includegraphics[width=0.24\linewidth]{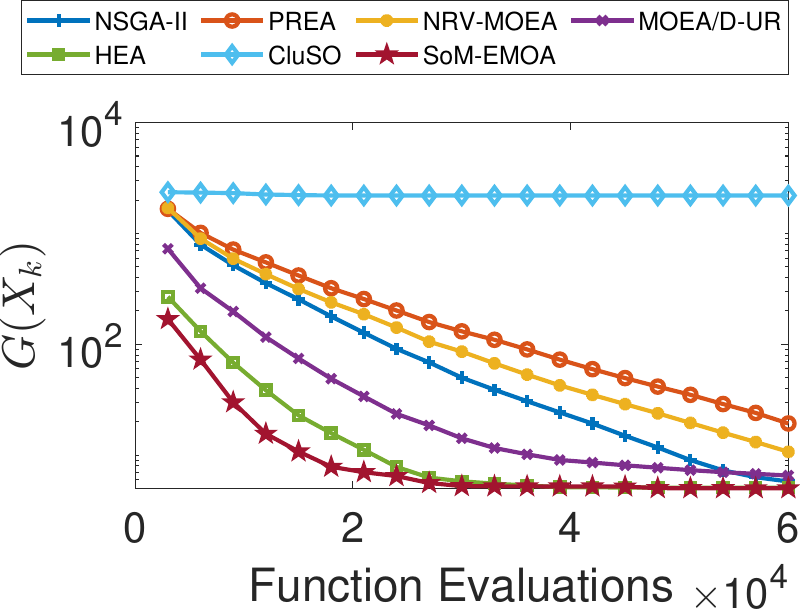} }
 \hspace{-8pt}\subfigure[F4M-DTLZ4]{ 
\includegraphics[width=0.24\linewidth]{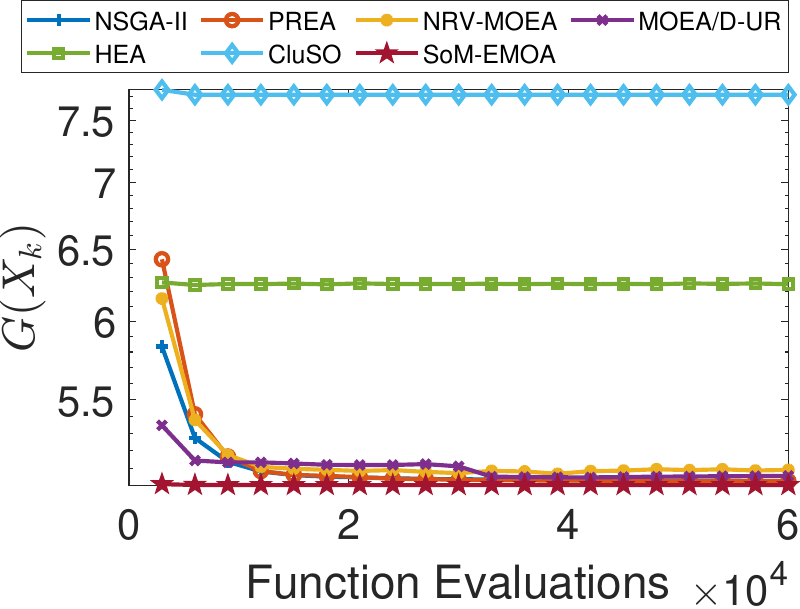} }
 \hspace{-8pt}\subfigure[F4M-WFG1]{ 
\includegraphics[width=0.24\linewidth]{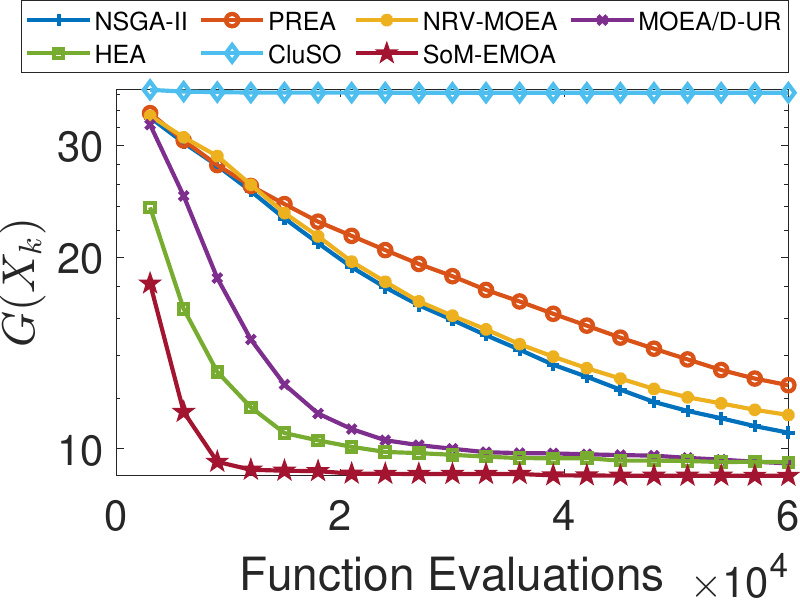} }
 \hspace{-8pt}\subfigure[F4M-WFG2]{ 
\includegraphics[width=0.24\linewidth]{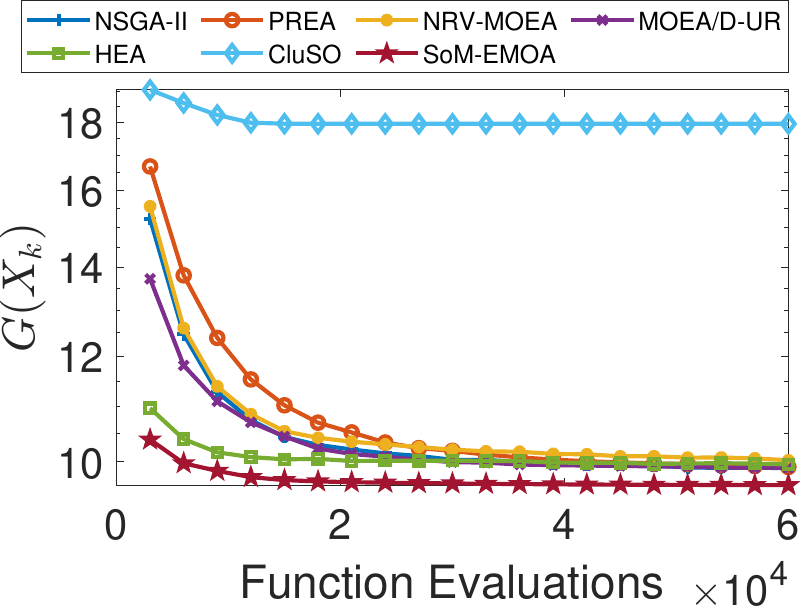} }
 \hspace{-8pt}\subfigure[F4M-WFG3]{ 
\includegraphics[width=0.24\linewidth]{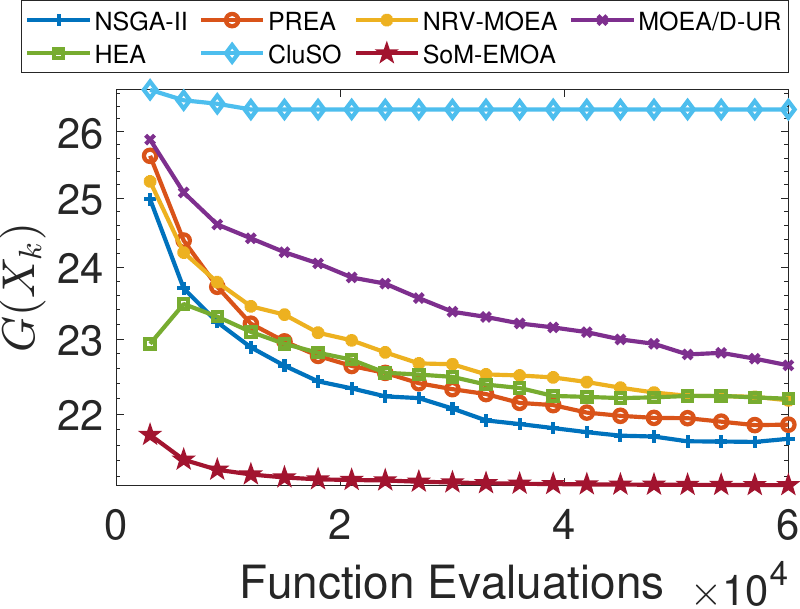} }
 \hspace{-8pt}\subfigure[F4M-WFG4]{ 
\includegraphics[width=0.24\linewidth]{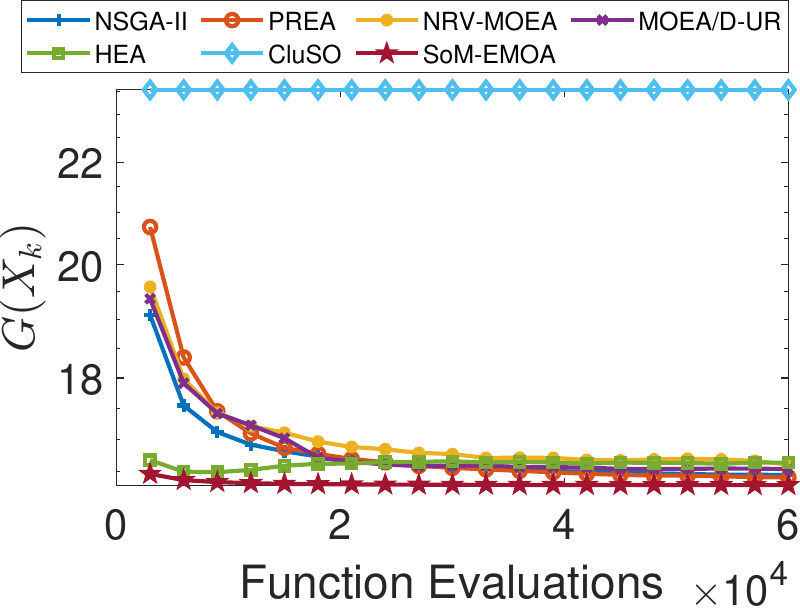} }
    \caption{{The convergence curve of different algorithms on the proposed F4M optimization benchmark problems. The number of objectives $m=50$ and the size of the solution set $k=5$.}}
    \label{fig:compare1}
\end{figure*}

Table~\ref{table:comapre1} presents the corresponding results on the proposed R2-based F4M benchmark suite for the case of $k=5$ (The results for $k=10$ are provided in the supplementary material). Once again, SoM-EMOA achieves the best performance in nearly all cases. Out of 32 test instances, it ranks best on 31 instances. This consistent superiority across a diverse range of benchmark problems verifies the strong generalizability and scalability of the proposed approach.

Fig.~\ref{fig:compare1} plots the convergence curves for the proposed benchmark suite when $m=50$ and $k=5$. SoM-EMOA achieves markedly lower $G_{\text{ws}}(X_k)$ values and faster convergence within the same number of function evaluations in most cases. In contrast, other algorithms either stagnate early or converge to inferior solution sets, indicating their limited capability in solving high-dimensional F4M problems.  We can observe that CluSO performs poorly on all test problems although it is specifically designed for solving F4M optimization problems. This shows that our proposed test suite is more challenging, which can stimulate further development of F4M optimization algorithms. 


{Another observation in Figure~\ref{fig:compare1} and Table~\ref{table:comapre1} is that some
traditional EMO algorithms such as NSGA-II and HEA
exhibit competitive performance on several F4M benchmark problems. 
Although these algorithms are not specifically designed for F4M optimization, 
their relatively good results can be explained by the nature of the constructed F4M test problems.}

{In our benchmark design, problems such as F4M-DTLZ and F4M-WFG are derived 
from standard three-objective DTLZ and WFG problems 
through the R2-based transformation described in Section~IV.
In this transformation, each scalarized F4M objective corresponds to 
a weighted direction in the original three-objective space, 
and every optimal solution for these scalarized objectives 
lies on the original Pareto front of the underlying MOP.
Consequently, algorithms like NSGA-II—which are known to approximate 
the Pareto front of three-objective problems effectively—can still 
generate populations distributed along the true front of the base problem.
When the F4M cover set is later constructed from these solutions, 
the resulting coverage quality can appear high,
even though the algorithm itself is not explicitly optimizing the F4M objective.}

\begin{table*}[!htb]
\renewcommand{\arraystretch}{1.2}
\centering
\caption{The mean values of $G_{\text{ws}}(X_k)$ obtained by SoM-EMOA and all baselines on the real-world test problems. The size of the final solution set $k$ is 5. The `$+$', `$-$' and `$\approx$' indicate that the compared algorithm is `significantly better than', `significantly worse than' and `statistically similar to' SoM-EMOA, respectively.}
\begin{tabular}{cccccccccc}
\toprule
Problem&FEs&NSGA-II&PREA&NRV-MOEA&MOEA/D-UR&HEA&CluSO&MOCOBO&SoM-EMOA\\
\midrule
\multirow{2}{*}{DDMOP1}&1000&-9.2519e+1 $-$&-8.6229e+1 $-$&-6.1651e+1 $-$&-1.0671e+2 $-$&-8.8777e+1 $-$&-1.0850e+2 $-$&-9.3182e+1 $-$&\hl{-1.2584e+2}\\
&10000&-1.2945e+2 $-$&-1.3024e+2 $-$&-1.0246e+2 $-$&-1.2763e+2 $-$&-1.0292e+2 $-$&-1.0786e+2 $-$&-&\hl{-1.3046e+2}\\
\hline
\multirow{2}{*}{DDMOP4}&1000&-4.4692e+3 $-$&-4.4662e+3 $-$&-4.4540e+3 $-$&-4.5150e+3 $-$&-4.4400e+3 $-$&\hl{-4.5348e+3 $+$}&-4.5238e+3 $-$&-4.5280e+3\\
&10000&-4.5350e+3 $-$&-4.5332e+3 $-$&-4.5070e+3 $-$&-4.5380e+3 $-$&-4.5115e+3 $-$&-4.5349e+3 $-$&-&\hl{-4.5389e+3}\\
\hline
\multicolumn{2}{c}{$+/-/\approx$}&0/4/0&0/4/0&0/4/0&0/4/0&0/4/0&1/3/0&0/2/0&-\\
\bottomrule
\label{table:compare2}
\end{tabular}
\end{table*}
{This phenomenon is particularly pronounced for F4M-DTLZ and F4M-WFG instances 
with $m$ derived from a low-dimensional base problem (e.g., $q=3$),
where the transformed objectives remain highly correlated.
In such cases, the original Pareto front inherently contains points 
that provide good coverage across the derived scalarized directions.
However, when the number of base objectives or the degree of correlation decreases,
the advantage of F4M-specific algorithms like SoM-EMOA and CluSO 
becomes more evident, as they explicitly optimize the set-level coverage objective.}

\subsection{Results on Real-world Problems}
\label{sec:realworld}

{Table~\ref{table:compare2} reports the mean values of  $G_{\text{ws}}(X_k)$ obtained by SoM-EMOA and seven  baseline algorithms 
on two real-world many-objective optimization problems, DDMOP1 and DDMOP4.
The results are presented under two different evaluation budgets, 
i.e., 1000 and 10\,000 function evaluations (FEs), 
with the final solution set size fixed at $k=5$.}

{Across both problems and evaluation budgets, SoM-EMOA consistently 
achieves the best or statistically comparable $G_{\text{ws}}(X_k)$ values. 
For DDMOP1, SoM-EMOA attains the lowest (best) coverage metric among all algorithms 
at both 1000 and 10\,000~FEs. 
These results indicate that SoM-EMOA can effectively discover 
a compact set of complementary solutions that jointly cover the diverse objectives 
even when the number of evaluations is very limited.}

{For the more complex DDMOP4 problem, the results exhibit a similar trend. 
SoM-EMOA performs comparably to CluSO at 1000~FEs 
and slightly better at 10\,000~FEs, achieving the best mean $G_{\text{ws}}(X_k)$ value. 
Notably, CluSO achieves a marginal advantage at 1000~FEs, 
suggesting that its clustering-based mechanism may offer a small initial exploration benefit.
However, as the number of evaluations increases, 
SoM-EMOA converges to the best performance, 
highlighting its superior search efficiency and robustness across generations.}

{For the MOCOBO algorithm, results are available only for the 1000-FE budget. 
Due to its surrogate-modeling and trust-region Bayesian optimization design,
MOCOBO becomes computationally prohibitive when the number of function evaluations 
is increased to 10\,000, and hence its results could not be obtained within reasonable time. 
Nevertheless, even under 1000~FEs, SoM-EMOA achieves significantly better performance 
than MOCOBO on both DDMOP1 and DDMOP4, 
demonstrating the advantage of the proposed evolutionary approach 
in scenarios where computational efficiency and limited evaluations are crucial.}


\subsection{Runtime Comparison}
\label{sec:runtime}

\begin{figure}[htbp]
    \centering
    \subfigure[Synthetic Problems]{ 
\includegraphics[width=0.8\linewidth]{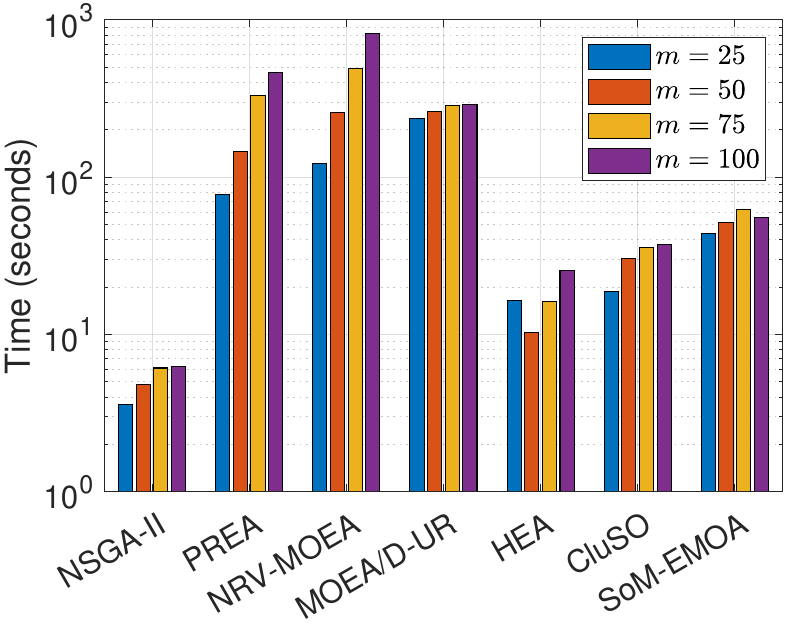} }           
    \subfigure[Real-world Problems]{ 
\includegraphics[width=0.8\linewidth]{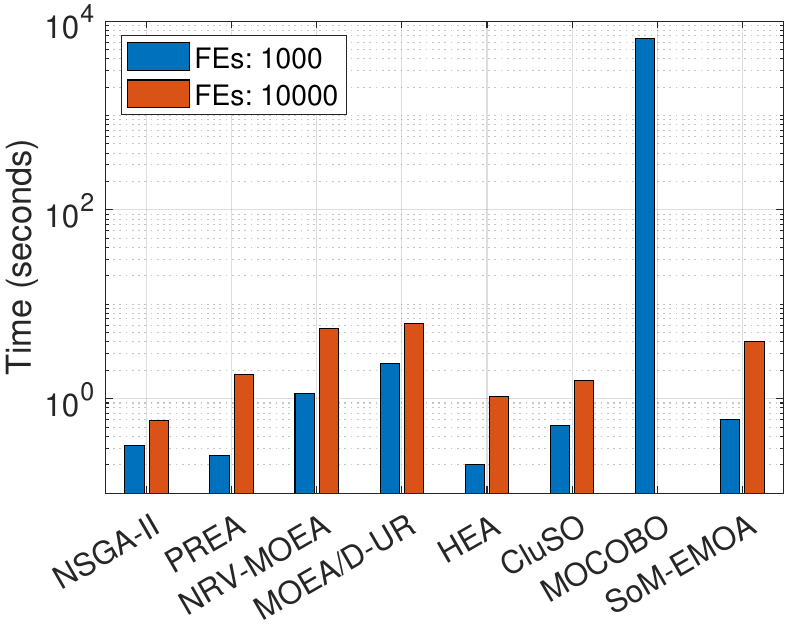} } 
\caption{{Comparison of runtime (in seconds) for different algorithms on synthetic and real-world problems. Each bar represents the average runtime of a single algorithm over all problems with $k=5$.}}
    \label{fig:runtime}
\end{figure}

{Fig.~\ref{fig:runtime} shows the runtime comparison of all algorithms 
on both the synthetic and real-world test problems. 
From these results, several observations can be made.}

{As shown in Fig.~\ref{fig:runtime}(a), 
NSGA-II achieves the lowest average runtime among all algorithms, 
followed closely by HEA and CluSO. 
In contrast, PREA, NRV-MOEA, and MOEA/D-UR 
incur higher computational costs due to their more complex 
decomposition and preference-learning components.
SoM-EMOA exhibits a moderate runtime overhead 
relative to the lightweight algorithms (e.g., NSGA-II and HEA),
which is expected since it performs additional 
set-level coverage evaluation and archive management operations.
Nevertheless, SoM-EMOA remains substantially faster 
than most preference-based and decomposition-based baselines.
MOCOBO is excluded from this comparison 
because its surrogate-modeling and trust-region updates 
require extensive computations; with the evaluation budget set to 60\,000 FEs, 
MOCOBO could not complete within a reasonable time.}

{The runtime comparison on real-world benchmarks is shown in Fig.~\ref{fig:runtime}(b).
When the evaluation budget is 1000 FEs, 
MOCOBO yields results but its runtime is orders of magnitude higher 
than all other algorithms, reflecting the heavy computational overhead 
of repeated Gaussian process training and trust-region optimization.
For the 10\,000-FE case, MOCOBO results are unavailable 
because its runtime becomes prohibitively long.
Among the remaining algorithms, NSGA-II again shows the fastest execution,
followed by HEA and CluSO, while SoM-EMOA maintains a moderate cost profile.
This indicates that the additional operations introduced in SoM-EMOA 
for coverage-based selection and archive updates 
do not substantially compromise its computational efficiency.}



\subsection{Ablation Studies}
\subsubsection{The Effectiveness of Archive $A$}
To investigate the effectiveness of the archive \( A \) in guiding the evolutionary process, we conduct an ablation study by comparing the original SoM-EMOA with a variant where the archive \( A \) is removed (denoted as SoM-EMOA-$\cancel{A}$). In this variant, offspring solutions are generated solely from the current population without utilizing the solutions in \( A \).

Table~\ref{table:ablation} presents the mean values of \( G_{\text{ws}}(X_k) \) for both SoM-EMOA and its archive-free variant SoM-EMOA-$\cancel{A}$ across 13 benchmark problems. The number of objectives is fixed at \( m = 50 \) and the size of the final solution set is \( k = 5 \). From the results, it is evident that the archive plays a critical role in enhancing performance. Specifically, SoM-EMOA outperforms its archive-free variant on 12 out of 13 problems, while achieving statistically similar performance on 1 problems. 

These results confirm the importance of maintaining archive \( A \) as a memory of high-quality, objective-specific representatives. By biasing offspring generation toward these archive solutions, the algorithm effectively ensures that all objectives are systematically addressed and that high-quality solutions are more likely to be discovered.

\subsubsection{The Effectiveness of Probability Distribution $\boldsymbol{p}$}
To evaluate the contribution of the probability distribution $\boldsymbol{p}$ in the mating selection process in Algorithm \ref{alg:mating-selection}, we compare the original SoM-EMOA with a variant in which $\boldsymbol{p}$ is not used (denoted as SoM-EMOA-$\cancel{\boldsymbol{p}}$). In this variant, the second parent solution is selected uniformly at random from the archive \( A \), without considering which objectives are poorly addressed by the current population.

As shown in Table~\ref{table:ablation}, the performance of SoM-EMOA is consistently superior to its variant SoM-EMOA-$\cancel{\boldsymbol{p}}$ on the majority of benchmark problems. Specifically, SoM-EMOA outperforms SoM-EMOA-$\cancel{\boldsymbol{p}}$ on 5 out of 13 problems, achieves statistically similar performance on 7 problems, and is only outperformed in 1 case. These results demonstrate that incorporating $\boldsymbol{p}$ provides a meaningful advantage.

This improvement stems from the adaptive nature of $\boldsymbol{p}$, which encodes the coverage deficiency of each objective. By probabilistically favoring archive solutions that target the most insufficiently addressed objectives, the mating process becomes more informed and goal-directed. Consequently, the use of $\boldsymbol{p}$ facilitates the discovery of more complementary solutions, ultimately leading to better coverage of all objectives in the final solution set.

\begin{table}[htbp]
\renewcommand{\arraystretch}{1.2}
\centering
\caption{The mean values of $G_{\text{ws}}(X_k)$ obtained by SoM-EMOA and its two variants (SoM-EMOA-$\cancel{A}$ indicates that archive $A$ is removed and SoM-EMOA-$\cancel{\boldsymbol{p}}$ indicates that probability distribution $\boldsymbol{p}$ is not used). The number of objectives $m$ is 50 and the size of the final solution set $k$ is 5. The `$+$', `$-$' and `$\approx$' indicate that the compared algorithm is `significantly better than', `significantly worse than' and `statistically similar to' SoM-EMOA, respectively.} 
\begin{tabular}{cccc}
\toprule
Problem&SoM-EMOA-$\cancel{A}$&SoM-EMOA-$\cancel{\boldsymbol{p}}$&SoM-EMOA\\
\midrule
\multirow{1}{*}{DC-MaTS1}&-2.7397e+1 $-$&-3.7724e+1 $-$&\hl{-3.9372e+1}\\
\hline
\multirow{1}{*}{DC-MaTS2}&-3.0596e+1 $-$&-3.9094e+1 $-$&\hl{-4.0098e+1}\\
\hline
\multirow{1}{*}{DC-MaTS3}&-2.5320e+1 $-$&-3.6205e+1 $-$&\hl{-3.8744e+1}\\
\hline
\multirow{1}{*}{DC-MaTS4}&-2.8279e+1 $-$&-3.8391e+1 $-$&\hl{-4.0054e+1}\\
\hline
\multirow{1}{*}{NMLR}&8.8613e-1 $-$&8.4252e-1 $\approx$&\hl{7.9513e-1}\\
\hline
\multirow{1}{*}{F4M-DTLZ1}&2.0644e+0 $-$&2.0627e+0 $\approx$&\hl{2.0601e+0}\\
\hline
\multirow{1}{*}{F4M-DTLZ2}&5.0129e+0 $-$&\hl{5.0048e+0 $\approx$}&5.0055e+0\\
\hline
\multirow{1}{*}{F4M-DTLZ3}&5.0461e+0 $-$&5.0249e+0 $\approx$&\hl{5.0234e+0}\\
\hline
\multirow{1}{*}{F4M-DTLZ4}&6.7652e+0 $-$&8.3773e+0 $-$&\hl{5.0043e+0}\\
\hline
\multirow{1}{*}{F4M-WFG1}&9.9371e+0 $-$&9.1178e+0 $\approx$&\hl{9.0636e+0}\\
\hline
\multirow{1}{*}{F4M-WFG2}&9.6334e+0 $-$&9.6107e+0 $\approx$&\hl{9.6015e+0}\\
\hline
\multirow{1}{*}{F4M-WFG3}&2.1577e+1 $-$&2.1159e+1 $\approx$&\hl{2.1094e+1}\\
\hline
\multirow{1}{*}{F4M-WFG4}&1.6294e+1 $\approx$&\hl{1.6292e+1 $+$}&1.6294e+1\\
\hline
\multicolumn{1}{c}{$+/-/\approx$}&0/12/1&1/5/7&\\
\bottomrule
\end{tabular}
\label{table:ablation}
\end{table}


\subsection{Limitations and Discussion}

{Although SoM-EMOA demonstrates competitive performance across a variety of F4M benchmark problems, 
it inherits several limitations that are intrinsic to stochastic evolutionary search.}


{The quality of the initial population and the choice of algorithmic parameters 
(e.g., mutation rate, archive size, and population size $k$) 
can noticeably influence the final coverage performance.
As SoM-EMOA is designed to operate with small populations ($k \ll m$), 
suboptimal initialization may limit its exploration ability in early generations. Two strategies can help mitigate stochastic variability:
\begin{itemize}
    \item \textbf{Hybridization with gradient-based search.}
    Incorporating gradient-based search (as in \cite{zhang2023predefined,lin2024few}) 
    could improve convergence speed and reduce the number of random evaluations.
    \item \textbf{Adaptive population control.}
    Dynamically adjusting the population or archive size based on coverage improvement 
    could stabilize convergence and avoid premature stagnation.
\end{itemize}}

{
In essence, the stochastic nature of SoM-EMOA offers flexibility in exploration 
but also introduces randomness that may affect reproducibility and consistency. 
Recognizing these limitations is crucial for future extensions that combine 
evolutionary search with adaptive, gradient-based, or deterministic components 
to achieve both robustness and efficiency in F4M optimization.}

\section{Conclusions}

In this paper, we proposed an efficient evolutionary algorithm for F4M optimization, which aims to find a small set of solutions that collectively address a large number of objectives. Unlike traditional EMO algorithms that seek a diverse approximation of the entire Pareto front, our method is tailored to the F4M's goal of coverage with minimal redundancy.

We also introduced a new R2-based benchmark test suite for F4M optimization. By leveraging the structural similarity between the F4M optimization objective and the R2 indicator, we constructed a flexible test suite that transforms standard MOPs into challenging F4M optimization instances. This overcomes the limitations of existing benchmarks such as DTLZ, WFG, and the overly simple DC-MaTS suite.

Experimental results on various F4M optimization instances with up to 100 objectives demonstrated the superior performance of our algorithm compared to  CluSO, MOCOBO, and several state-of-the-art EMO algorithms. The proposed algorithm consistently achieved better performance on a wide variety of test problems.


{In future work, the proposed SoM-EMOA framework could be extended to online and zeroth-order optimization settings. 
For instance, an online variant could incorporate event-triggered updates and time-varying archives to handle streaming or non-stationary objectives, 
inspired by developments in distributed bandit convex optimization with time-varying constraints~\cite{zhang2025distributed,zhang2025one}. 
Moreover, SoM-EMOA is inherently compatible with zeroth-order (bandit) feedback since it relies only on function evaluations; 
integrating one-point or stochastic perturbation techniques could enable more sample-efficient exploration in noisy or expensive environments. 
Such extensions would further broaden the applicability of F4M optimization to real-time and data-limited scenarios.}

\bibliographystyle{IEEEtran}
\bibliography{sample-bibliography} 

\end{document}